%% file: evocua.tex
\documentclass{article}

\usepackage{wrapfig}
\usepackage{CJKutf8} 
\usepackage{tabularx} 
\usepackage{evocua_style}

\usepackage{xcolor}     
\usepackage{tcolorbox}  

\usepackage{float}
\usepackage{algorithm}
\usepackage{algpseudocode}

\usepackage{pgfplots}
\pgfplotsset{compat=1.18}

\usepackage[noblocks]{authblk}

\makeatletter
\renewcommand\AB@affilsepx{ \quad } 
\makeatother

\usepackage{graphicx}
\usepackage{booktabs}
\usepackage{listings}
\usepackage{xcolor}
\usepackage{subcaption}
\usepackage{geometry}
\definecolor{codegray}{rgb}{0.5,0.5,0.5}
\definecolor{codepurple}{rgb}{0.58,0,0.82}
\definecolor{backcolour}{rgb}{0.95,0.95,0.92}
\lstdefinestyle{jsonstyle}{
    backgroundcolor=\color{backcolour},
    commentstyle=\color{codegray},
    keywordstyle=\color{magenta},
    stringstyle=\color{codepurple},
    basicstyle=\ttfamily\footnotesize,
    breakatwhitespace=false,
    breaklines=true,
    captionpos=b,
    keepspaces=true,
    showspaces=false,
    showstringspaces=false,
    showtabs=false,
    tabsize=2,
    frame=single
}

\usepackage[utf8]{inputenc} 
\usepackage[T1]{fontenc}    
\usepackage{newunicodechar}
\newunicodechar{，}{,}
\usepackage{hyperref}       
\usepackage{xcolor}
\usepackage{svg} 

\hypersetup{
    colorlinks=true,      
    linkcolor=blue,      
    urlcolor=blue,       
    citecolor=blue,      
    linkbordercolor=blue, 
    urlbordercolor=blue,
    citebordercolor=blue,
    pdfborderstyle={/S/U/W 1}, 
}

\usepackage{graphicx}
\usepackage{subcaption}
\usepackage{tcolorbox}
\usepackage{listings}
\usepackage{xcolor}
\usepackage{fontawesome5} 

\definecolor{codegreen}{rgb}{0,0.6,0}
\definecolor{codegray}{rgb}{0.5,0.5,0.5}
\definecolor{codepurple}{rgb}{0.58,0,0.82}
\definecolor{backcolour}{rgb}{0.95,0.95,0.92}

\lstdefinelanguage{json}{
    basicstyle=\ttfamily\footnotesize,
    numbers=none,
    backgroundcolor=\color{backcolour},
    stringstyle=\color{codepurple},
    keywordstyle=\color{blue},
    showstringspaces=false,
    breaklines=true,
    frame=single,
    rulecolor=\color{lightgray},
}

\usepackage{xurl}            
\usepackage{booktabs}       
\usepackage{amsfonts}       
\usepackage{nicefrac}       
\usepackage{microtype}      
\usepackage{lipsum}		
\usepackage{graphicx}
\usepackage{natbib}
\usepackage{doi}
\usepackage{amsmath}
\usepackage{amssymb} 
\usepackage{xspace}
\usepackage{enumitem}
\usepackage{multirow}
\usepackage{multicol}
\usepackage{subcaption} 
\usepackage{makecell}
\usepackage{hyperref, cleveref}
\setlist[itemize]{leftmargin=*}
\setlist[enumerate]{leftmargin=*}
\setlist[description]{leftmargin=*}

\usepackage{colortbl}
\definecolor{mygray}{gray}{.88}
\definecolor{mycyan}{cmyk}{.15,0,0,0}
\definecolor{mycyan2}{cmyk}{.85,0,0,0}

\definecolor{mygreen}{rgb}{0.19, 0.79, 0.02}
\definecolor{midnightgreen}{rgb}{0.0, 0.29, 0.33}

\definecolor{codegray}{rgb}{0.5,0.5,0.5}
\definecolor{codepurple}{rgb}{0.58,0,0.82}
\definecolor{backcolour}{rgb}{0.95,0.95,0.92}

\title{EvoCUA: Evolving Computer Use Agents via Learning from Scalable Synthetic Experience}

\author[1]{Taofeng Xue\textsuperscript{*,$\dagger$}}
\author[1]{Chong Peng\textsuperscript{*,$\dagger$}}
\author[1,2]{Mianqiu Huang\textsuperscript{*}}

\author[1]{Linsen Guo}
\author[1,3]{Tiancheng Han}
\author[1,4]{Haozhe Wang}
\author[1]{\protect\\Jianing Wang}
\author[1]{Xiaocheng Zhang}
\author[1]{Xin Yang}
\author[1]{Dengchang Zhao}
\author[1]{Jinrui Ding}
\author[1]{Xiandi Ma}
\author[1]{\protect\\Yuchen Xie}
\author[1]{Peng Pei}
\author[1]{Xunliang Cai}
\author[2]{Xipeng Qiu}

\affil[1]{Meituan}
\affil[2]{ Fudan University}
\affil[3]{Tongji University}
\affil[4]{The Hong Kong University of Science and Technology}

\renewcommand{\headeright}{}

\renewcommand{\headeright}{\raisebox{-0.2\height}{\includegraphics[width=5em]{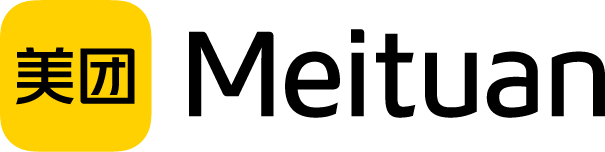}}}

\renewcommand{\shorttitle}{EvoCUA: Evolving Computer Use Agents via Learning from Scalable Synthetic Experience}

\begin{document}
\maketitle

{
    \renewcommand{\thefootnote}{}
    \footnotetext{
        \textsuperscript{*}Equal contribution. \quad
        \textsuperscript{$\dagger$}Corresponding authors.
    }
}

\vspace{-0.8cm}

\input{sections/abstract}


\newpage

\input{sections/introduction}

\input{sections/formulation}

\input{sections/data}

\input{sections/system}

\input{sections/algorithm}

\input{sections/evaluation}

\input{sections/onlinerl}

\input{sections/related_work}

\input{sections/conclusion}

\input{sections/acknowledgements}

\newpage

\bibliographystyle{unsrtnat}
\bibliography{references}

\newpage
\appendix

\input{sections/appendix}

\end{document}

%% file: sections/abstract.tex
\begin{abstract}
The development of native computer-use agents (CUA) represents a significant leap in multimodal AI. 
However, their potential is currently bottlenecked by the constraints of static data scaling. 
Existing paradigms relying primarily on passive imitation of static datasets struggle to capture the intricate causal dynamics inherent in long-horizon computer tasks. 
In this work, we introduce EvoCUA, a native computer use agentic model.
Unlike static imitation, 
EvoCUA integrates data generation and policy optimization into a self-sustaining evolutionary cycle. 
To mitigate data scarcity, we develop a verifiable synthesis engine that autonomously generates diverse tasks coupled with executable validators. 
To enable large-scale experience acquisition, we design a scalable infrastructure orchestrating tens of thousands of asynchronous sandbox rollouts. 
Building on these massive trajectories, we propose an iterative evolving learning strategy to efficiently internalize this experience. 
This mechanism dynamically regulates policy updates by identifying capability boundaries—reinforcing successful routines while transforming failure trajectories into rich supervision through error analysis and self-correction.
Empirical evaluations on the OSWorld benchmark demonstrate that EvoCUA achieves a success rate of 56.7\%, establishing a new open-source state-of-the-art. Notably, EvoCUA significantly outperforms the previous best open-source model, OpenCUA-72B (45.0\%), and surpasses leading closed-weights models such as UI-TARS-2 (53.1\%).
Crucially, our results underscore the generalizability of this approach: the evolving paradigm driven by learning from experience yields consistent performance gains across foundation models of varying scales, establishing a robust and scalable path for advancing native agent capabilities.

{ \footnotesize
  \setlength{\parskip}{0.2pt} 
  \noindent \textbf{Github}: \url{https://github.com/meituan/EvoCUA} \\
  \noindent \textbf{Huggingface}: \url{https://huggingface.co/meituan/EvoCUA-32B-20260105} \\
  \noindent \textbf{OSWorld}: \url{https://os-world.github.io/}
}

\end{abstract}


\begin{figure}[H]
    \centering
\includegraphics[width=0.7\linewidth]{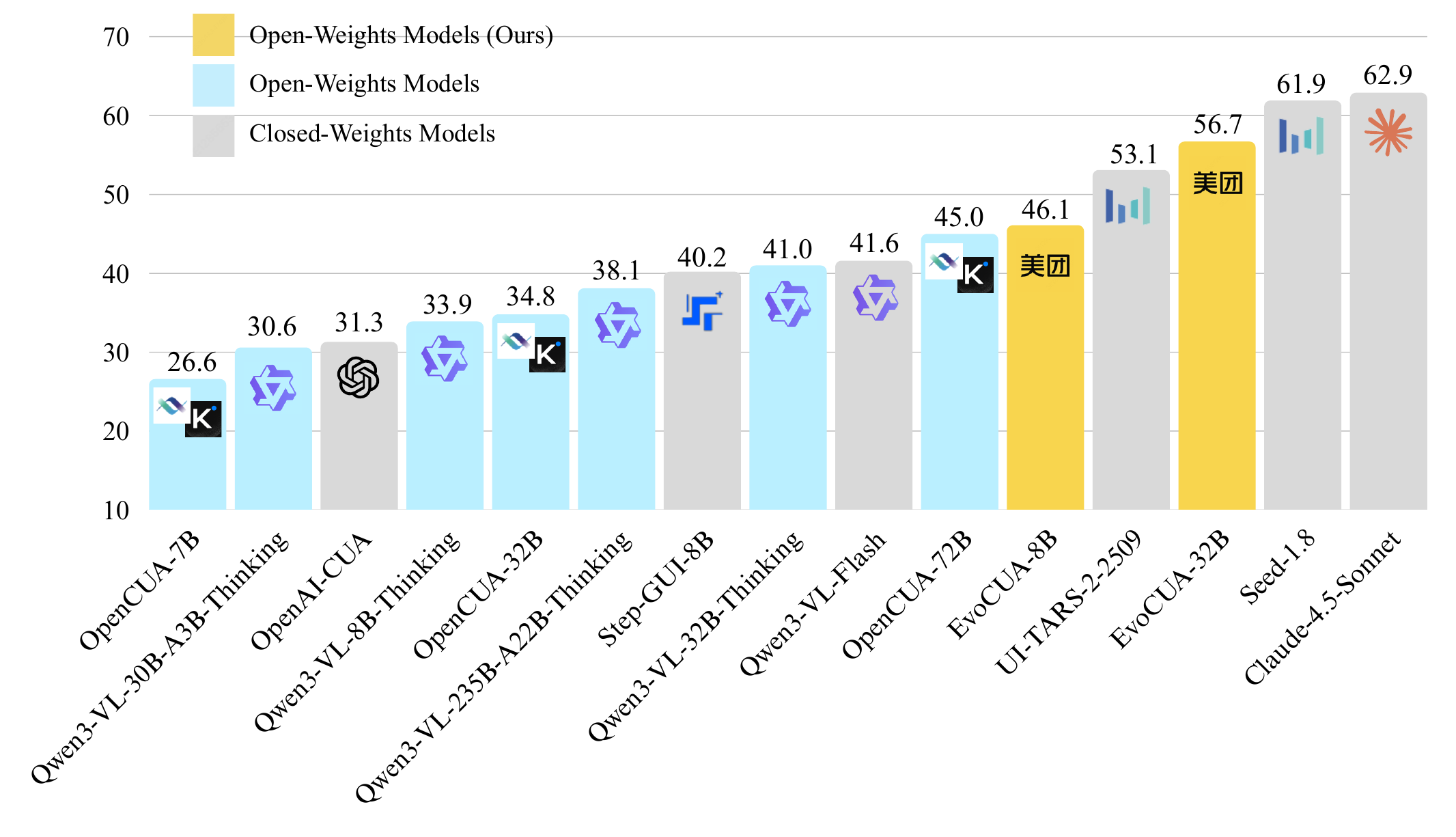}
    \caption{\textbf{Performance comparison on the OSWorld-Verified benchmark.} 
    Our \textbf{EvoCUA-32B} achieves state-of-the-art performance (56.7\%) among open-weights models.}
    \label{fig:osworld_perf}
\end{figure}



%% file: sections/introduction.tex
\section{Introduction}

The development of generalist agents capable of mastering Graphical User Interfaces (GUIs) represents a pivotal milestone toward artificial general intelligence. 
Unlike specialized tools, these agents must perceive complex visual contexts and execute long-horizon workflows across heterogeneous applications, effectively emulating human-computer interaction. 
While recent native vision-language models (VLMs) have successfully integrated perception and action into end-to-end architectures~\citep{Qwen3-VL, bytedance2025seed18}, achieving human-level reliability remains a significant challenge. 
Despite the foundational architectures established by state-of-the-art efforts such as UI-TARS-2~\citep{wang2025ui}, and OpenCUA~\citep{wang2025opencua}, further progress is increasingly constrained by a critical bottleneck: the diminishing returns of scaling with static datasets.


Existing scaling laws are largely confined to passive imitation of fixed, non-interactive datasets, failing to capture the causal feedback inherent in real-world computer use. Overcoming this limitation necessitates a paradigm shift from data scaling via static traces to experience scaling via massive interactive rollouts. Dynamic experience provides a richer supervisory signal than static text, encompassing environmental feedback and critical insights from both success and failure. 
However, transforming raw interaction into a self-improving learning loop presents three primary challenges:
1) \emph{Verifiable data synthesis}. Merely synthesizing textual queries often leads to hallucinations, where the agent generates plausible plans for infeasible tasks. Consequently, a robust framework is essential to ensure that generated queries are strictly grounded in solvable states, aligning with the principles of verifiable rewards. 
2) \emph{Scalable interaction infrastructure}: High-throughput experience production demands a unified system that integrates massive environment simulation with high-performance reinforcement learning to support continuous, asynchronous interaction. 
3) \emph{Efficient training recipe}:
Given an large-scale interaction space, unconstrained exploration is computationally prohibitive. Effective learning requires an on-policy approach that mimics human learning dynamics: consolidating mastered routines while focusing intensely on boundary tasks where the agent oscillates between success and failure.

\begin{figure*}[t]
\centering
\includegraphics[width=\textwidth]{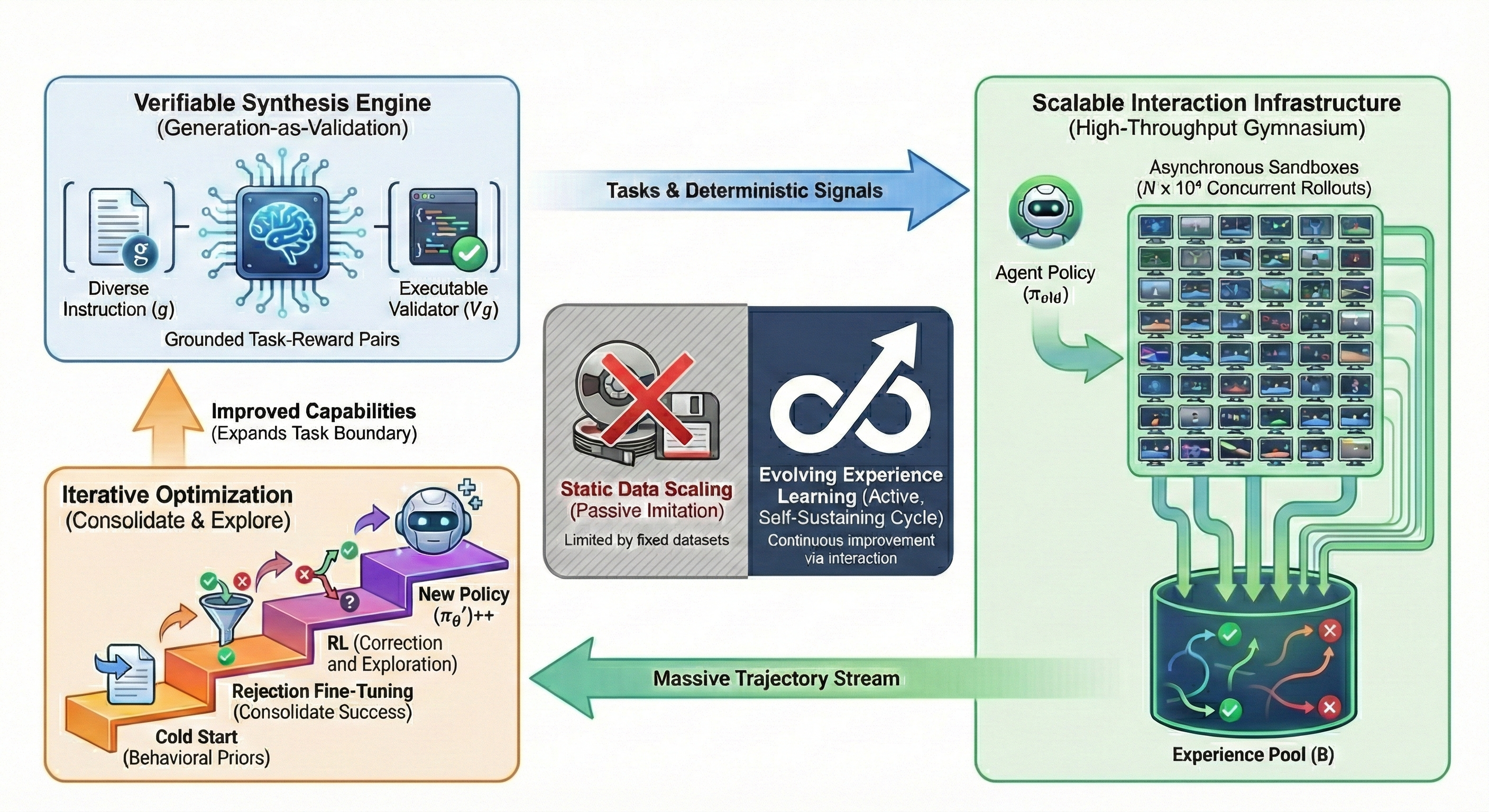}
\caption{Overview of \textbf{EvoCUA}. The diagram illustrates the paradigm shift from static imitation to an active evolving experience learning cycle (center). The approach unifies three core modules: the \textbf{Verifiable Synthesis Engine} (top left); the \textbf{Scalable Interaction Infrastructure} (right); and \textbf{Iterative Optimization} (bottom left).}
\label{fig:evocua_overall}
\end{figure*}

To address these issues, in this report, we introduce EvoCUA, a native computer use agent that addresses these challenges through the evolving paradigm driven by learning from experience. 
As illustrated in Figure \ref{fig:evocua_overall}, by unifying verifiable synthesis, high-throughput infrastructure, and evolutionary optimization, 
EvoCUA establishes a self-sustaining cycle that continuously transforms synthetic compute into high-quality agent capabilities. 
Our core contributions are threefold:

\begin{itemize}

\item \textbf{Verifiable Synthesis Engine}. To overcome the data bottleneck while ensuring strict environmental grounding, we first propose a synthesis engine that autonomously generates diverse tasks alongside their executable validators. 
Moving beyond text-only generation, we analyze atomic capabilities to synthesize self-contained task definitions. 
This "Generation-as-Validation" approach eliminates the ambiguity of natural language rewards, providing the agent with precise, deterministic supervision signals.

\item \textbf{Scalable Interaction Infrastructure}. To support the magnitude of experience scaling required, we construct a high-performance infrastructure that integrates a massive sandbox environment. 
Beyond mere trajectory generation, this system functions as a dynamic gymnasium, providing the real-time feedback and state transitions essential for on-policy optimization. 
By architecting a fully asynchronous rollout mechanism, we decouple simulation from model updates, enabling the system to orchestrate tens of thousands of concurrent interactive sessions.

\item \textbf{Evolving Paradigm via Learning from Experience}. We introduce an iterative training paradigm centered on learning from experience to ensure efficiency. The process begins with a diversity-aware cold start to establish robust priors. Subsequently, through continuous environmental exploration, the model contrasts successful versus failed trajectories to consolidate effective patterns and rectify errors. This dynamic feedback loop transforms accumulated experience into model parameters, yielding a precise and robust execution policy.

\end{itemize}

Empirical evaluations demonstrate that EvoCUA achieves a state-of-the-art success rate of 56.7\% on the OSWorld benchmark~\citep{xie2024osworld}, significantly outperforming the previous open-source SOTA, OpenCUA-72B (45.0\%)~\citep{wang2025opencua}, and surpassing leading closed-source models UI-TARS-2(53.1\%)~\citep{wang2025ui}. Furthermore, the evolving experience learning paradigm proves to be a generalizable path, yielding consistent gains across multiple foundation models of varying sizes.

%% file: sections/formulation.tex
\section{Preliminaries}

Before introducing our EvoCUA, we provide the basic task definition of CUA in the following.
Formally, CUA can be viewed as a Partially Observable Markov Decision Process (POMDP)~\citep{kaelbling1998planning} with explicit reasoning, which is optimized through a co-evolutionary cycle of verifiable task synthesis and policy refinement.

\subsection{POMDP}
Given a natural language instruction $g$, the interaction process is modeled as a tuple $(\mathcal{S}, \mathcal{A}, \mathcal{Z}, \mathcal{O}, \mathcal{P}, \mathcal{R}_{syn})$, where $\mathcal{S}$, $\mathcal{A}$, $\mathcal{Z}$, $\mathcal{O}$, $\mathcal{P}$, and $\mathcal{R}_{syn}$ denotes to the state space, action space, thought space, observation, transition kernel and reward function, respectively. The details are shown in the following:

\begin{itemize}

\item{\textbf{State Space} ($\mathcal{S}$)}: The environment is modeled with an underlying computer system state $s_t \in \mathcal{S}$, which includes application states, system configurations, and implicit system-level context. This state is not directly observable by the agent. Instead, the agent perceives a visual observation rendered from the state, $ I_t \triangleq \mathrm{Render}(s_t) \in \mathbb{R}^{H \times W \times 3} $, corresponding to the screen image at time $t$. $H, W$ denote the height and width size of the screenshot, respectively. The rendered screenshot $I_t$ serves as the sole perceptual interface through which the agent observes the environment.

\item \textbf{Observation ($\mathcal{O}$)}: 
At step $t$, the agent receives a raw visual observation $o_t \in \mathcal{O}$, where $o_t \triangleq I_t \in \mathbb{R}^{H \times W \times 3}$.
To address partial observability, we define the interaction history
$h_t = \{ g, o_0, z_0, a_0, \dots, o_{t-1}, z_{t-1}, a_{t-1} \},$
which serves as the conditioning context for the agent's decision-making process.
In practical implementations, to prevent the context window from being flooded, we perform context engineering strategies following \citep{wang2025opencua, Qwen3-VL}. We restrict the visual history to the five most recent screenshots and compress the textual history using a structured inner monologue with action representation to balance performance and token efficiency. 

\item \textbf{Action Space ($\mathcal{A}$)}: We define a unified native action space $\mathcal{A}$ that 
encompasses coordinate-based mouse events $\mathcal{A}_{\text{mouse}}$, keyboard inputs $\mathcal{A}_{\text{keyboard}}$, and special control  $\mathcal{A}_{\text{control}}$ primitives for managing the task execution flow.
Formally, we defined $\mathcal{A} = \mathcal{A}_{\text{mouse}} \cup \mathcal{A}_{\text{keyboard}} \cup \mathcal{A}_{\text{control}}$.

\item \textbf{Thought Space ($\mathcal{Z}$)}: We explicitly model the reasoning process as a internal thought space $\mathcal{Z}$. At each step $t$, the agent generates a natural language reasoning trace $z_t \in \mathcal{Z}$ before acting.
It serves as an intermediate cognitive state internal to the agent, used to ground the subsequent physical action in the current visual context.

\item \textbf{Policy ($\pi_\theta$)}: 
The agent follows a parameterized policy $\pi_\theta(z_t, a_t \mid h_t, o_t)$ that governs both reasoning and action selection. At each step $t$, the policy first generates a reasoning trace $z_t$ conditioned on the current interaction context, and subsequently selects an executable action $a_t$ conditioned on the generated reasoning. 
This sequential generation ensures that action execution is conditional on explicit reasoning.

\item \textbf{Transition ($\mathcal{P}$)}: 
The environment state evolves according to a state transition kernel $\mathcal{P}(s_{t+1} \mid s_t, a_t)$, which captures the dynamics of the underlying computer system in response to the executed physical action $a_t$. Given the updated state $s_{t+1}$, the subsequent visual observation is rendered as $I_{t+1} = \mathrm{Render}(s_{t+1})$.

\item \textbf{Verifiable Reward ($\mathcal{R}_{syn}$)}: 
Supervision is grounded in execution correctness via a verifiable synthesis mechanism. For a given instruction $g$, the synthesis engine provides an executable validator $V_g$ that evaluates whether the task objective is satisfied. We define a sparse, binary, instruction-conditioned reward based on the terminal environment state:
$
\mathcal{R}_{syn}(s_T; g) \triangleq \mathbb{I}\!\left[V_g(s_T) = \mathrm{True}\right],
$
where $s_T$ denotes the environment state at episode termination. This reward formulation provides outcome-level supervision without requiring intermediate annotations.

\end{itemize}

\subsection{Objective}
Rather than viewing the training data as a static dataset, we conceptualize it as a dynamic distribution that 
is adaptively parameterized conditioned on the current policy snapshot $\pi_{\text{old}}$.
The optimization objective $J(\theta)$ is formulated to maximize the verification rate over a coupled curriculum orchestrated by the synthesis engine $\mathcal{T}_{syn}$:

\begin{itemize}

\item \textbf{Theoretical Objective:} Formally, our goal is to maximize the expected success rate over a distribution of tasks that evolves adaptively based on the current policy's capability ($\pi_{\text{old}}$):$$J(\theta) = \mathbb{E}_{(g, V_g) \sim \mathcal{T}_{syn}(\cdot | \pi_{\text{old}})} \left[ \mathbb{E}_{\tau \sim \pi_\theta(\cdot|g)} [\mathcal{R}_{syn}(s_T;g)] \right],$$
where $\mathcal{T}_{syn}(\cdot | \pi_{\text{old}})$ represents the synthesis engine's distribution, which dynamically adjusts task complexity and diversity based on the agent's performance. We use $\tau \sim \pi_\theta(\cdot \mid g)$ to denote trajectories induced by executing policy $\pi_\theta$ in the environment dynamics $\mathcal{P}$ under instruction $g$.

\item \textbf{Empirical Approximation:}
As the expectation above does not admit a closed-form solution, we resort to an empirical approximation via massive-scale Monte Carlo estimation. 
The scalable interaction infrastructure maintains a transient Experience Pool $\mathcal{B}$ that aggregates a high-throughput stream of fresh interaction trajectories:
$$
\mathcal{B} = \{ (\tau, V_g) \mid \tau \sim \pi_{\text{old}}(\cdot|g), \ (g, V_g) \sim \mathcal{T}_{syn} \},
$$
where $\pi_{\text{old}}$ denotes the policy snapshots driving tens of thousands of asynchronous sandboxes. By continuously updating $\theta$ using batches sampled from $\mathcal{B}$, we effectively close the loop between verifiable synthesis, large-scale execution, and on-policy optimization.

\end{itemize}


Building upon this formulation, the following sections detail the implementation of the three core pillars of EvoCUA. Section 3 introduces the Verifiable Synthesis Engine, detailing the generation of the coupled distribution $(g, V_g)$.
Section 4 describes the Scalable Interaction Gymnasium, the infrastructure that facilitates the massive-scale rollout pool $\mathcal{B}$.
Finally, Section 5 elaborates on the Evolving Paradigm via Learning from Experience, demonstrating the initialization and iterative optimization of 
$\pi_\theta$ to achieve state-of-the-art performance.

%% file: sections/data.tex
\section{Verifiable Synthesis Engine}
\label{sec:verifiable_synthesis}

\begin{figure}[htbp]
    \centering
    \vspace{-0.8cm}
\includegraphics[width=1.0\linewidth]{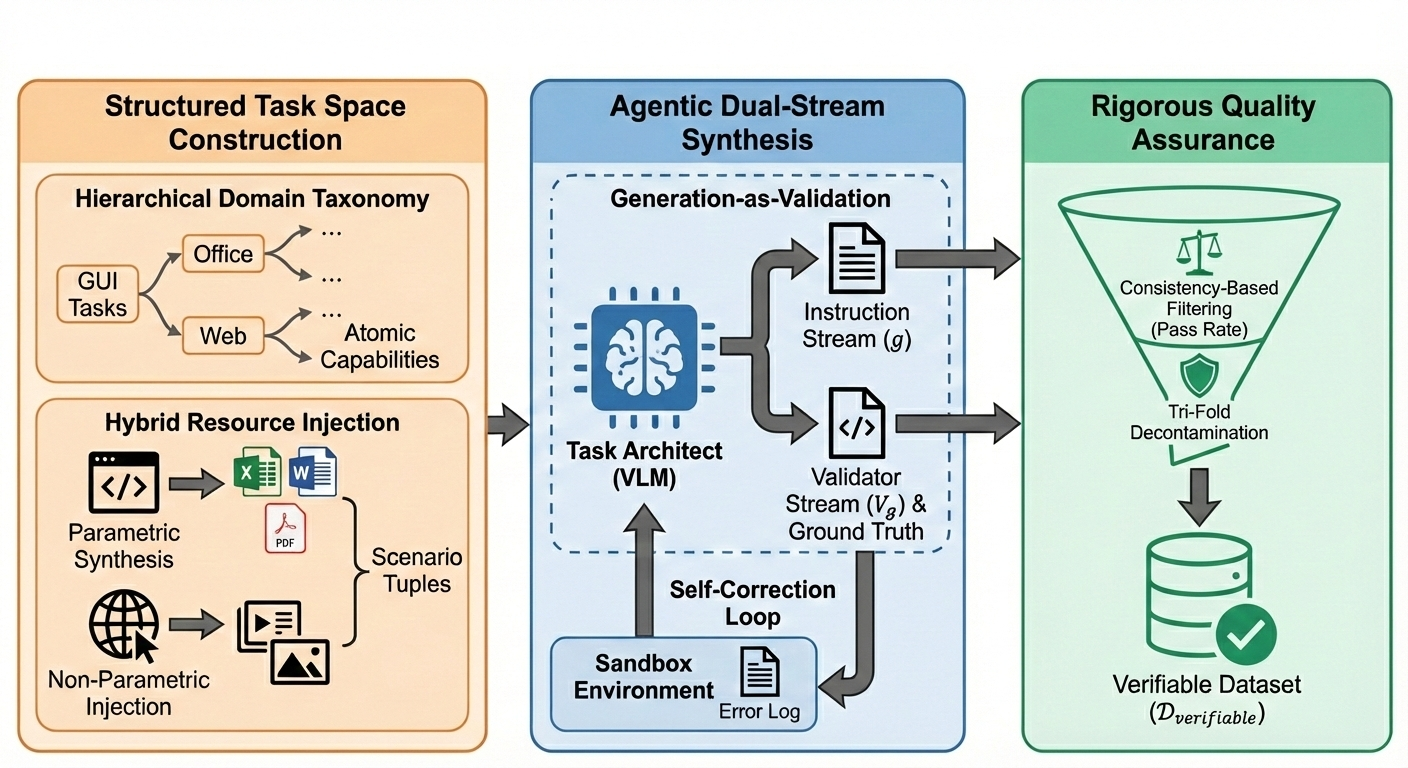}
    \caption{Architecture of the Verifiable Synthesis Engine. The pipeline operates in three cascading stages: (1) \textbf{Structured Task Space Construction} to define diverse scenarios from domain taxonomies and hybrid resources; (2) \textbf{Agentic Dual-Stream Synthesis}, where a Task Architect (VLM) co-generates instructions ($g$) and executable validators ($V_g$) via a closed-loop feedback mechanism; and (3) \textbf{Rigorous Quality Assurance} to filter outputs for high consistency and ensure decontamination, yielding the final verifiable dataset.}
    \label{fig:vse_architecture}
\end{figure}

In this section, we introduce a Verifiable Synthesis Engine, which focuses on overcoming the inherent limitations, such as reward hacking, and the absence of precise training signals.
Unlike passive data collection, 
Based on this engine, we can implement the operation on the ``generation-as-validation'' paradigm , which is illustrated in Figure \ref{fig:vse_architecture}.

Formally, 
given a synthesized instruction,
$g$, the engine must co-generate a deterministic, executable validator $V_g$. 
This ensures that the reward signal $\mathcal{R}_{syn}(s_T;g)$ 
is derived from a strict verification of the final environment state, thereby bypassing the ambiguity of semantic matching
The architecture is organized into three cascading modules: structured task space construction, agentic dual-stream synthesis, and rigorous quality assurance.

\subsection{Structured Task Space Construction}
To ensure the synthesized distribution $\mathcal{T}_{syn}$ captures the complexity of real-world computer use, we first establish a structured task space decomposed into domains and resources.

\paragraph{Hierarchical Domain Taxonomy.}
We argue that atomic capabilities are inherently transferable and compositionally form complex tasks. Guided by this principle, we systematically categorize core desktop applications (e.g., Web Browsers, Excel, and Word) and decompose user behaviors into \textit{atomic capabilities}. This orthogonal decomposition enables the agent to generalize to diverse scenarios through the recombination of primitive skills. For instance, a financial analysis task in Excel is decomposed into sub-skills such as formula manipulation, data sorting, and chart generation. Leveraging a hierarchical domain taxonomy, we synthesized a wide range of task scenarios featuring diverse user personas~\citep{ge2024scaling} to ensure data diversity. Synthesized scenarios range from educators designing lecture slides to algorithm engineers conducting technical literature surveys.

\paragraph{Hybrid Resource Injection.}
To bridge the simulation-to-reality gap, we implement a hybrid strategy for the environment's initial state:
\begin{itemize}
    \item \textit{Parametric synthesis:} For structured data (e.g., production sales data), we utilize code-based generators to batch-produce documents (Word, Excel, PDF) by parameterizing variables such as names, prices and dates. This ensures high variability in numerical values and layouts.
    \item \textit{Non-parametric injection:} To mitigate the sterility of synthetic templates, we inject public internet data (e.g., images, audio, complex slides). This forces the agent to handle the visual noise and structural diversity inherent in real-world files.
\end{itemize}

\subsection{Agentic Dual-Stream Synthesis}
The core synthesis process is modeled as a ReAct-based agentic workflow~\citep{yao2022react}. Given a sampled scenario tuple (Role, Capability, Resources), a foundation VLM functions as a \textit{task architect} to execute a dual-stream generation:

\begin{enumerate}
    \item \textit{Instruction stream ($g$):} The architect formulates a natural language query grounded in the specific resource context, ensuring user intent is clear and achievable.
    \item \textit{Validator stream ($V_g$):} Simultaneously, the architect generates the ground truth (GT) and the corresponding executable evaluator code. This code defines the precise success conditions for the task~\citep{yang2025ultracua}.
\end{enumerate}

To guarantee executability, we enforce a \textit{closed-loop feedback mechanism}. The generated code is immediately executed in a real sandbox environment. The execution results — including output files from successful runs, as well as error messages from failed executions (e.g., syntax errors, API mismatches) — are fed back to the model to evaluate the quality of GT files and the evaluator. This process iterates multiple rounds until the execution succeeds and passes quality checks. To further enhance stability, we abstract frequently used verification logic into a standardized tool library. Finally, the valid tuple is formatted into a standardized JSON structure compatible with established benchmarks like OSWorld.

\subsection{Rigorous Quality Assurance}
The final stage filters the raw synthesized pairs $\{(g, V_g)\}$ through a rigorous protocol to eliminate false positives (hallucinated success), false negative and data leakage.

\paragraph{Consistency-based filtering.}
We deploy a reference computer use agent to perform sandbox rollouts on the synthesized tasks. We enforce a high bar for data inclusion. First, tasks that fail to complete the rollout due to issues such as parameter configuration anomalies will return error messages to the ReAct-based agentic workflow for modification. Second, for tasks with successful rollouts, we calculate pass rates using both a reward model and an evaluator. Organized by our hierarchical domain taxonomy, we perform manual spot checks on tasks where the pass rates from these two sources show significant discrepancies. For cases where manual inspection identifies clear evaluator failures leading to false positives or false negatives, we refine the ReAct-based agentic workflow to mitigate these issues. Finally, we preserve tasks that are cross-verified by the sandbox rollout, the reward model, and manual inspection.

\paragraph{Tri-fold decontamination.}
While synthetic data generation effectively mitigates the scarcity of high-quality trajectories, it introduces the risk of data leakage, as powerful models may inadvertently reproduce benchmark content from their vast pre-training corpora. To prevent inflated metrics and ensure the validity of our experimental insights, we enforce a rigorous decontamination: (1) \textit{semantic decontamination}, using LLM-based filtering to remove instructions semantically equivalent to benchmark queries; (2) \textit{configuration decontamination}, pruning tasks with identical application initialization settings within certain domains; and (3) \textit{evaluator decontamination}, verifying that the generated success conditions and ground truth files do not overlap with existing evaluation scripts.

Through this pipeline, we have successfully scaled verifiable training data to tens of thousands of instances, effectively breaking the bottleneck of manual data curation.

%% file: sections/system.tex
\section{Scalable Interaction Infrastructure}

\begin{figure}[htbp]
    \centering
    \includegraphics[width=\linewidth]{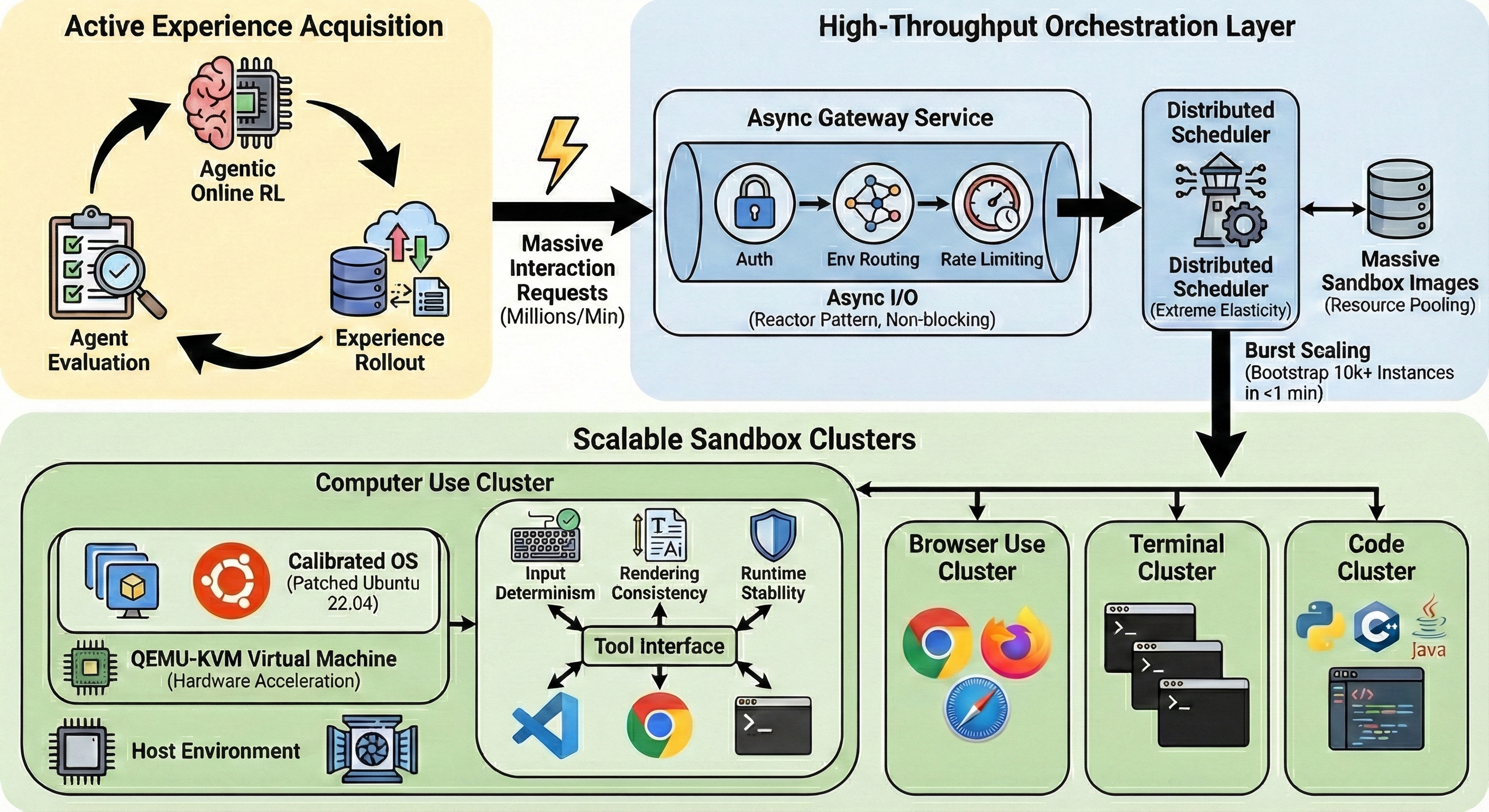}

    \caption{\textbf{Scalable Infrastructure.} The architecture orchestrates massive interaction requests from the online RL loop (top-left) through an asynchronous gateway and distributed scheduler (top-right). The bottom layer deploys parallel sandbox clusters, highlighting the \textbf{Computer Use Sandbox}, which utilizes QEMU-KVM virtualization and a calibrated OS to ensure input determinism, rendering consistency, and runtime stability for high-fidelity environments.}
    
    \label{fig:system_architecture}
\end{figure}

The transition from static data scaling to evolving experience learning necessitates a fundamental shift in infrastructure capabilities. Unlike passive training pipelines, our active learning paradigm requires a high-throughput gymnasium capable of generating continuous, diverse, and interactive feedback at a massive scale. To address the challenges of heterogeneity, high concurrency, and strict session isolation inherent in large-scale reinforcement learning, we developed a unified environment sandbox platform. This platform,  illustrated in Figure \ref{fig:system_architecture}, serves as the bedrock for EvoCUA, orchestrating hundreds of thousands of daily sandbox sessions and processing millions of interaction requests per day with industrial-grade stability.

\subsection{Architecture and Abstractions}
To manage the complexity of diverse interaction tasks, the platform is architected around two core abstractions: \textit{tools} and \textit{clusters}.

\paragraph{Tools.} A tool encapsulates the immutable definition of a simulation environment, including version-controlled system images and exposed interaction APIs. The platform currently supports hundreds of distinct environment types, ranging from generic benchmarks to specialized agentic environments. This design decouples environment iteration from experimentation, ensuring backward compatibility and reproducibility.

\paragraph{Clusters (Dynamic Scaling Units).} A cluster represents the runtime instantiation of a tool and serves as the fundamental unit for environment scaling. By specifying tool types and configuring resource quotas, users can instantly provision customized environment services for distinct workloads. This abstraction allows the infrastructure to dynamically scale environment instances—from a handful of debugging sessions to tens of thousands of concurrent training nodes—without resource contention or cross-contamination.

\subsection{High-Throughput Orchestration}
The capability to support massive-scale exploration hinges on the efficiency of our microservices architecture, specifically designed to eliminate I/O bottlenecks and enable rapid environment scaling.

The infrastructure relies on an asynchronous gateway service based on the reactor pattern for non-blocking I/O. This service achieves a routing throughput at the scale of hundreds of thousands of requests per minute. By decoupling the control plane (lifecycle management) from the data plane (environment interaction), the gateway prevents long-running environment executions from blocking critical routing logic.

Complementing the gateway, the distributed scheduler is engineered for extreme elasticity, managing the lifecycle of massive sandbox images. Leveraging distributed sharding and resource pooling, the scheduler achieves high-efficiency node scheduling. More critically, it supports burst scaling capabilities, bootstrapping tens of thousands of sandbox instances within one minute. This rapid instantiation ensures that the environment scaling strictly matches the training demand of on-policy reinforcement learning, minimizing the latency between policy updates and experience collection. Ultimately, this resilient scheduling backbone enables the infrastructure to stably sustain over 100,000 concurrent sandboxes.

\subsection{High-Fidelity Environment Instantiation}
To support the rigorous requirements of computer use tasks, we implement a hybrid virtualization architecture that encapsulates QEMU-KVM virtual machines within Docker containers.

\paragraph{Hybrid virtualization.} While Docker provides compatibility with our orchestration layer, the internal execution relies on QEMU with KVM hardware acceleration. We construct a customized QEMU launch sequence that explicitly disables non-essential peripherals while optimizing I/O performance. This nested design ensures strict kernel-level isolation—crucial for security when agents execute arbitrary code—while maintaining near-native performance for GUI rendering and I/O operations.

\paragraph{Deterministic environment calibration.} We constructed a customized OS image based on Ubuntu 22.04 to address the gap between simulation and real-world deployment, implementing specific kernel and userspace patches:
\begin{itemize}
    \item \textbf{Input determinism (HID patching):} Standard virtualization often suffers from key mapping collisions. We calibrated the human interface device mapping at the \texttt{xkb} kernel level. Specifically, we modified the \texttt{/usr/share/x11/xkb/symbols/pc} definitions to resolve symbolic collisions (e.g., the \texttt{<} vs \texttt{>} shift-state error in US layouts), ensuring that the agent's symbolic intent strictly matches the realized character input.
    \item \textbf{Rendering consistency:} To prevent layout shifts in office software that confuse visual agents, we injected a comprehensive suite of proprietary fonts directly into the system font cache (\texttt{fc-cache}). This guarantees that documents render identically to their native counterparts.
    \item \textbf{Runtime stability:} The image is hardened with system-level proxy configurations to resolve network instabilities and includes pre-installed dependencies like \texttt{xsel} and \texttt{qpdf} to eliminate common runtime errors during clipboard operations and PDF processing.
\end{itemize}

%% file: sections/algorithm.tex
\section{Evolving Paradigm via Learning from Experience}

To bridge the gap between atomic imitation and generalist problem-solving, we propose the evolving paradigm via learning from experience. This paradigm shifts from static data scaling to a dynamic capability evolution cycle. The process is structured into three progressive stages: a supervised cold-start to establish behavioral priors, rejection sampling fine-tuning to consolidate successful experiences via adaptive scaling, and reinforcement learning to rectify failures and explore complex dynamics through interaction.

\subsection{Cold-Start}
\label{sec:prior_injection}

To initialize the policy $\pi_{\text{init}}$ with a robust behavioral prior, we construct a dataset $\mathcal{D}_{\text{prior}}$ containing trajectories that exhibit both precise execution and coherent reasoning. We first formally define the unified action and thought spaces to establish the structural bounds of the agent, and subsequently leverage these definitions to synthesize and format grounded interaction data.

\vspace{0.5em}
\noindent \textbf{Unifying the Action Space ($\mathcal{A}$).}
We implement \textbf{Semantic Action Mapping} to construct a unified action space $\mathcal{A} = \mathcal{A}_{\text{mouse}} \cup \mathcal{A}_{\text{keyboard}} \cup \mathcal{A}_{\text{control}}$ as illustrated in Appendix \ref{appendix:action_space}. We categorize raw event streams into two primary components:
\begin{itemize}
    \item \textit{Physical Interaction ($\mathcal{A}_{\text{mouse}} \cup \mathcal{A}_{\text{keyboard}}$)}: This component encompasses coordinate-based mouse events and keyboard inputs. Crucially, to support complex, multi-step operations, we implement a \textbf{Stateful Interaction} mechanism. By decoupling discrete key presses into \texttt{key\_down} and \texttt{key\_up} events, the policy can maintain active states (e.g., holding modifiers like \texttt{Shift} for multi-selection) required for complex tasks.
    \item \textit{Control Primitives ($\mathcal{A}_{\text{control}}$)}: We introduce meta-actions to manage the execution flow distinct from physical I/O. Specifically, the \texttt{wait} primitive allows the agent to handle asynchronous UI rendering, while \texttt{terminate} serves as a formal signal to conclude the task.
\end{itemize}

\vspace{0.5em}
\noindent \textbf{Structuring the Thought Space ($\mathcal{Z}$).}
To enable interpretable and robust decision-making, we define a \textbf{Reasoning Schema} for the latent thought space $\mathcal{Z}$. This schema imposes a structured format to ensure the reasoning process strictly aligns with execution logic:
\begin{itemize}
    \item \textit{Goal Clarification ($z_{0}$)}: At the initial step ($t=0$), the agent is required to explicitly paraphrase the user's objective. This clarifies ambiguous instructions and grounds the subsequent planning process.
    \item \textit{Observation Consistency ($z_{\text{obs}}$)}: To minimize hallucinations, the reasoning trace must include a concise summary of key visual elements. We enforce strict semantic consistency between this textual summary and the actual observed state.
    \item \textit{Self-Verification ($z_{\text{check}}$)}: Before issuing the final termination signal, the agent is prompted to execute auxiliary interaction steps (e.g., checking a file status) to visually confirm that the execution result aligns with the user's instruction.
    \item \textit{Reflection and Correction ($z_{\text{reflect}}$)}: We leverage failed rollouts for error correction. Upon identifying a critical error step in a failed trajectory, we restore the environment to the pre-error state. To account for sandbox non-determinism, we strictly filter for \textbf{state consistency} between the restored environment and the original trace. From this valid restored state, we induce self-correction using high-temperature sampling to generate successful remedial paths.
    \item \textit{Reasoning-Augmented Termination ($z_{T}$)}: To prevent the model from overfitting to the termination label, the \texttt{terminate} action must be strictly conditional on a preceding reasoning trace. This trace requires the agent to explicitly synthesize visual evidence to justify task completion, ensuring the decision is grounded in logic rather than memorized patterns.
\end{itemize}

\vspace{0.5em}
Based on these formalized definitions, we synthesize the prior dataset $\mathcal{D}_{\text{prior}}$ by leveraging foundational vision-language models (e.g., Qwen3-VL, OpenCUA) within a modular framework. Crucially, to ensure alignment between reasoning and action, we employ a \textbf{Hindsight Reasoning Generation} strategy. Treating the ground-truth execution path as known future information, we retrospectively generate reasoning traces $z_t$ that explain the observed actions, thereby augmenting physical trajectories with coherent cognitive chains.

\textbf{Training Details}. For model training, we decompose these multi-turn trajectories into single-turn samples. To balance information density with memory constraints, the input context retains full multimodal details (screenshots, reasoning, and actions) only for the most recent five steps, while earlier history is compressed into text-only semantic actions. The training loss is computed exclusively on the current step's reasoning and action.

Finally, to preserve general foundation capabilities, we incorporate a diverse mixture of general-purpose data, covering STEM, OCR, visual grounding, and text-based reasoning. The volume of this general data is balanced to match the scale of the decomposed single-turn trajectory samples.

\textbf{Qualitative Analysis}. We synthesize trajectory data adhering to this schema. Following cold start training, qualitative analysis confirms the agent effectively masters atomic capabilities as illustrated in Appendix \ref{appendix:vis_tool}. However, a critical robustness gap persists in complex scenarios. While the agent can execute standard long-horizon workflows, it exhibits fragility in boundary cases. To address these limitations, we move to the next stage: internalizing scalable, high-quality experiences.

\subsection{Rejection Sampling Fine-Tuning}
\label{sec:rft}

The objective of Rejection Sampling Fine-Tuning (RFT)~\citep{ahn2024large} is to consolidate the agent's ability to solve tasks by learning exclusively from high-quality, successful executions. This process involves two key components: efficiently generating successful trajectories via dynamic compute, and denoising them to maximize the signal-to-noise ratio.

\textbf{Dynamic Compute Budgeting}. To optimize the generation of high-quality experience under computational constraints, we propose dynamic compute budgeting. Instead of uniformly allocating rollout resources, this mechanism adapts the exploration budget to the agent's current proficiency level for each specific task.

We establish a hierarchical budget spectrum $\mathcal{K} = \{k_1, \dots, k_n\}$ paired with descending success rate thresholds $\Lambda = \{\tau_1, \dots, \tau_n\}$. For a given task query $g$ drawn from the synthesis engine $\mathcal{T}_{\text{syn}}$, the system identifies the optimal rollout budget $K^*$ that satisfies the sufficiency condition:
\begin{align}
K^* = k_{i^*} \quad \text{where} \quad i^* = \min \{ i \mid \text{SR}(k_i) \ge \tau_i \}
\end{align}
Here, $\text{SR}(k_i)$ represents the pass rate observed with budget $k_i$. This strategy effectively prunes efficiently solved tasks and concentrates computational power on boundary queries—tasks where the policy exhibits high variance.

\textbf{Step-Level Denoising}. Although successful rollouts demonstrate the model's capability, they often contain significant noise. We use a judge model to analyze the trajectories and mask out redundant steps. This filtering is especially important for infeasible tasks; for these, we remove all intermediate actions and strictly keep the reasoning trace and the final \texttt{terminate=failure} action. This process refines the raw data into high-quality supervision, which is then aggregated into the experience pool $\mathcal{B}$.

Through this generation and filtering pipeline, we scale our high-fidelity experience pool $\mathcal{B}$ to tens of thousands of trajectories. We interleave this domain-specific experience with a balanced corpus of general-purpose multimodal data to prevent catastrophic forgetting.

\subsection{Reinforcement Learning}
While RFT consolidates what the agent \textit{can} do, it does not explicitly correct what it \textit{does wrong}. To push the capability boundary, we employ RL to learn from failures and explore via online interaction.

\begin{figure}[t]
\centering
\includegraphics[width=0.6\linewidth]{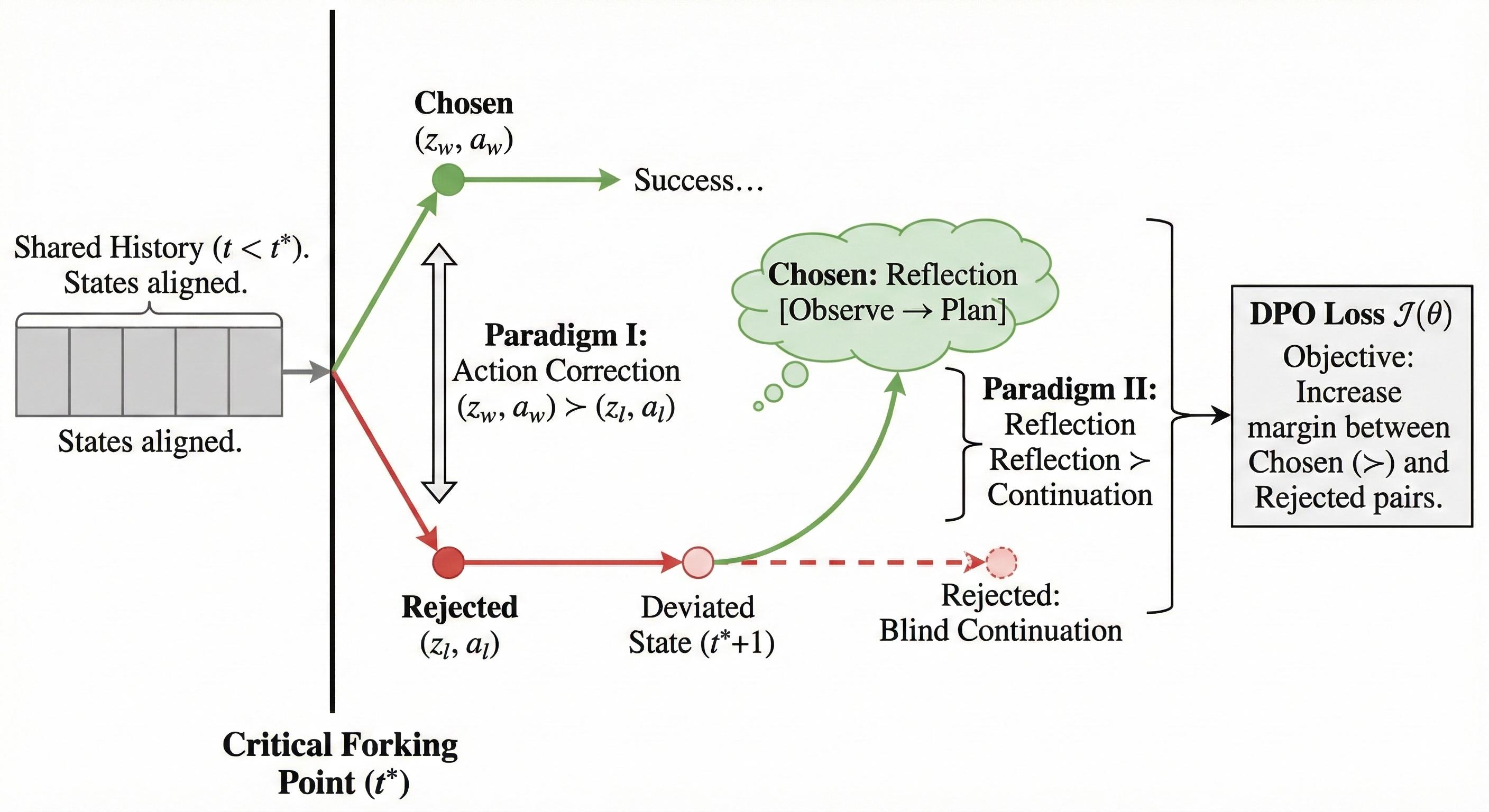}
\caption{Overview of the Dual-Paradigm DPO.
The process begins at a critical forking point $t^*$. Paradigm I (Action Correction) establishes a preference for the chosen action $(z_w, a_w)$ over the rejected action $(z_l, a_l)$. Paradigm II (Reflection) addresses the deviated state at $t^*+1$, prioritizing Reflection over Blind Continuation. Both paradigms define preference pairs that optimize the DPO Loss $\mathcal{J}(\theta)$ to maximize the margin between effective and ineffective strategies.}
\label{fig:evocua_dpo}
\end{figure}

Standard trajectory-level preference optimization is ill-suited for long-horizon tasks due to state misalignment. We instead propose a Step-Level Direct Preference Optimization strategy~\citep{lai2024step} that targets \textit{Critical Forking Points} illustrated in Figure \ref{fig:evocua_dpo}.

\textbf{Causal Deviation Discovery}. Given a failed rollout $\tau^-$ and a successful reference $\tau^+$ (retrieved from the same or a semantically equivalent task), we employ a Reference-Guided Diagnosis mechanism. We identify the Critical Deviation Step $t^*$ as the first timestamp where the agent's action diverges from the reference, despite the environmental states remaining functionally equivalent. This isolates the specific response $(z_{t^*}^-, a_{t^*}^-)$ that caused the agent to leave the optimal solution manifold.

\textbf{Structured Preference Construction}. Once the critical error $(z_l, a_l) = (z_{t^*}^-, a_{t^*}^-)$ is identified, we construct preference pairs to provide comprehensive supervision.

\begin{itemize}
 
\item \textbf{Paradigm I: Action Correction} (At Step $t^*$). The objective is to replace the rejected error $(z_l, a_l)$ with an optimal chosen response $(z_w, a_w)$. We obtain $(z_w, a_w)$ via window-based reference alignment (migrating thoughts and actions from $\tau^+$ via VLM semantic matching) or visual-grounded synthesis (synthesizing fresh traces via a general model when no alignment exists).

\item \textbf{Paradigm II: Reflection and Recovery} (At Step $t^*+1$). To improve robustness, we address the state immediately \textit{after} the error ($t^*+1$). We treat the agent's blind continuation as the rejected sample. For the chosen sample, we synthesize a Reflection Trace. Instead of acting blindly, the agent is trained to halt and generate a reasoning chain that: (1) observes the unexpected screen state, and (2) formulates a remedial plan.
\end{itemize}

\textbf{Optimization Objective}. We optimize the policy $\pi_\theta$ using Direct Preference Optimization (DPO). Consistent with our formulation where the policy generates a reasoning trace $z$ and an action $a$ conditioned on history $h_t$ and observation $o_t$, the loss function is defined as:
\begin{align}
\tiny
\mathcal{J}(\theta) = - \mathbb{E}_{(h_t, o_t, (z, a)_w, (z, a)_l) \sim \mathcal{D}}\left[\log \sigma \left( \beta \log \frac{\pi_\theta(z_w, a_w|h_t, o_t)}{\pi_{\text{ref}}(z_w, a_w|h_t, o_t)} - \beta \log \frac{\pi_\theta(z_l, a_l|h_t, o_t)}{\pi_{\text{ref}}(z_l, a_l|h_t, o_t)} \right) \right].
\end{align}
By iteratively updating the policy with these structured preferences, EvoCUA continuously expands its capability boundary, effectively converting transient interaction experience into robust model parameters.

In summary, the evolving experience learning paradigm establishes a rigorous cycle for enhancing agent reliability. By synergizing rejection fine-tuning to consolidate fundamental execution patterns with reinforcement learning to rectify errors in complex, long-tail scenarios, EvoCUA iteratively transforms scalable synthetic experience into policy parameters. This dual mechanism ensures that the agent not only stabilizes performance on standard tasks but also significantly improves robustness and generalization across boundary conditions, thereby realizing a more stable and universal computer use capability.

%% file: sections/evaluation.tex
\section{Evaluation}
\label{sec:evaluation}
In this section, we conduct a comprehensive empirical evaluation of EvoCUA. Our analysis focuses on three critical dimensions: (1) \textbf{Online Agentic Capability}, assessing long-horizon interaction in realistic environments; (2) \textbf{Offline Grounding}, evaluating fine-grained UI element understanding; and (3) \textbf{General VLM Capabilities}, ensuring the preservation of general multimodal reasoning.

\subsection{Experimental Setup}
\label{subsec:setup}

To advance beyond static imitation, we adopt a unified training process that begins with a lightweight cold start phase, utilizing approximately 1k high-quality trajectories to establish the complete action space and the structured reasoning pattern. Subsequently, the model enters a continuous iterative optimization cycle that combines experience generation with policy refinement. In this evolving phase, we progressively expand the training distribution by collecting successful trajectories from large-scale rejection sampling, applying step-level denoising, while simultaneously optimizing the policy through a mix of preference learning derived from errors and online exploration in realistic environments. This entire process is driven by a pass@k-guided dynamic compute strategy, which automatically focuses computational resources on harder queries and synthesizes supplementary data for under-performing domains, ensuring continuous capability growth across iterations.

We validate our approach across varying scales by post-training on the \textbf{Qwen3-VL-Thinking} \citep{Qwen3-VL} (8B, 32B) and \textbf{OpenCUA} \citep{wang2025opencua} (7B, 32B, 72B) foundation models.

\subsection{Main Results}
\label{subsec:main_results}

\subsubsection{Online Agent Evaluation}
We evaluate EvoCUA on the \textbf{OSWorld} benchmark, which serves as a representative testbed for open-ended computer use tasks. As summarized in Table \ref{tab:main_results}, our results highlight the effectiveness of the proposed method:

\begin{itemize}
    \item \textbf{State-of-the-Art Open-Weights Performance.} Our primary model, \textbf{EvoCUA-32B}, fine-tuned from the Qwen3-VL-32B-Thinking \citep{Qwen3-VL} backbone, achieves a success rate of \textbf{56.7\%}. This performance secures the top rank among all evaluated open-weights models.
    
    \item \textbf{Significant Improvements \& Efficiency.} EvoCUA-32B demonstrates a \textbf{+11.7\%} absolute improvement over the previous state-of-the-art open model, OpenCUA-72B (45.0\%), and a \textbf{+15.1\%} gain over its base model. Notably, these results are achieved under a strict 50-step constraint, whereas baselines typically require a 100-step budget to reach peak performance, indicating our model's superior execution precision.

    \item \textbf{Competitive with Closed-Weights Frontiers.} EvoCUA-32B effectively closes the gap with closed-weights models. Most notably, it outperforms the strong closed-weights baseline UI-TARS-2-2509 (53.1\%) by a margin of +3.6\%. Under equivalent step constraints, the performance gap between EvoCUA-32B and the industry-leading Claude-4.5-Sonnet (58.1\%) is narrowed to a mere \textbf{1.4\%}.

    \item \textbf{Scaling Efficiency \& Training Superiority.} The efficacy of our approach extends to smaller model scales. \textbf{EvoCUA-8B} achieves a success rate of \textbf{46.1\%}, surpassing specialized 72B-parameter models such as OpenCUA-72B. A direct comparison with Step-GUI-8B \citep{yan2025step} is particularly illuminating: although both models are initialized from the identical Qwen3-VL-8B backbone, EvoCUA-8B achieves a \textbf{+5.9\%} higher success rate (46.1\% vs. 40.2\%). This strictly isolates the contribution of our evolving experience learning paradigm, confirming that our data synthesis and RL strategies unlock significantly greater potential from the same foundational architecture.
    
\end{itemize}

\begin{table*}[t]
\centering
\caption{\textbf{Performance comparison on the OSWorld-Verified benchmark.} Models are categorized by accessibility (Closed-Weights vs. Open-Weights). \textbf{Max Steps} denotes the interaction budget per task. \textbf{EvoCUA-32B} achieves state-of-the-art performance among open models, significantly outperforming larger baselines.}
\label{tab:main_results}
\resizebox{0.8\textwidth}{!}{%
\begin{tabular}{llccc}
\toprule
\textbf{Model} & \textbf{Type} & \textbf{Max Steps} & \textbf{Success Rate (Pass@1)} \\
\midrule
\multicolumn{4}{l}{\textit{\textbf{Closed-Weights Models}}} \\
OpenAI CUA \citep{openai2025cua} & Specialized & 50 & 31.3\% \\
Step-GUI-8B \citep{yan2025step} & Specialized & 100 & 40.2\% \\
Qwen3-VL-Flash \citep{Qwen3-VL} & General & 100 & 41.6\% \\
UI-TARS-2-2509 \citep{wang2025ui} & General & 100 & 53.1\% \\
Claude-4.5-Sonnet \citep{anthropic2025claude45} & General & 50 & 58.1\% \\
Seed-1.8 \citep{bytedance2025seed18} & General & 100 & 61.9\% \\
Claude-4.5-Sonnet \citep{anthropic2025claude45} & General & 100 & \textbf{62.9\%} \\
\midrule
\multicolumn{4}{l}{\textit{\textbf{Open-Weights Models}}} \\
Qwen2.5-VL-32B-Instruct \citep{Qwen2.5-VL} & General & 100 & 5.9\% \\
Qwen2.5-VL-72B-Instruct \citep{Qwen2.5-VL} & General & 100 & 8.8\% \\
ScaleCUA-32B \citep{liu2025scalecua} & Specialized & 50 & 17.7\% \\
UI-TARS-72B-DPO \citep{qin2025ui} & Specialized & 50 & 24.6\% \\
OpenCUA-7B \citep{wang2025opencua} & Specialized & 100 & 26.6\% \\
UI-TARS-1.5-7B \citep{qin2025ui} & Specialized & 100 & 27.5\% \\
Qwen3-VL-8B-Thinking \citep{Qwen3-VL} & General & 100 & 30.6\% \\
OpenCUA-32B \citep{wang2025opencua} & Specialized & 100 & 34.8\% \\
GUI-Owl-7B-Desktop-RL \citep{ye2025mobile} & Specialized & 15 & 34.9\% \\
Qwen3-VL-235B-A22B Thinking \citep{Qwen3-VL} & General & 100 & 38.1\% \\
Qwen3-VL-32B-Thinking \citep{Qwen3-VL} & General & 100 & 41.0\% \\ 
OpenCUA-72B \citep{wang2025opencua} & Specialized & 100 & 45.0\% \\
\textbf{EvoCUA-8B (Ours)} & \textbf{General} & \textbf{50} & \textbf{46.1\%} \\
\textbf{EvoCUA-32B (Ours)} & \textbf{General} & \textbf{50} & \textbf{56.7\%} \\ 
\bottomrule
\end{tabular}%
}
\end{table*}

\subsubsection{Offline Grounding and General Capabilities}
\label{subsec:combined_eval}

We assess EvoCUA's performance across two critical dimensions: fine-grained GUI grounding (ScreenSpot-v2~\citep{wu2024atlas}, ScreenSpot-Pro~\citep{li2025screenspotpro}, OSWorld-G~\citep{xie2025scalingcomputerusegroundinguser}) and general multimodal robustness (MMMU~\citep{yue2024mmmumassivemultidisciplinemultimodal}, MMMU-Pro~\citep{yue2025mmmuprorobustmultidisciplinemultimodal}, MathVista~\citep{lu2024mathvistaevaluatingmathematicalreasoning}, MMStar~\citep{chen2024rightwayevaluatinglarge}, OCRBench~\citep{Liu_2024}). Table \ref{tab:combined_results} summarizes the results across different model scales and backbones.

\textbf{Analysis.} 
We observe distinct behaviors depending on the base model used. For the OpenCUA-72B backbone, our post-training strategy maintains performance parity or yields slight improvements across both grounding and general benchmarks (e.g., preserving MMMU scores while improving OSWorld-G). This stability confirms that our training method effectively preserves the base model's knowledge when the data distribution is aligned.

Conversely, the EvoCUA-32B variant exhibits performance decline in specific metrics, notably on ScreenSpot-Pro and MMMU, compared to the Qwen3-VL-32B-Thinking baseline. We attribute this performance drop primarily to discrepancies in data distribution and patterns. Due to time constraints, the general dataset used for fine-tuning EvoCUA was directly adopted from OpenCUA-72B variants experiments. However, this dataset is non-thinking, creating a significant mismatch with the thinking-based distribution of the Qwen3-VL-32B-Thinking model. We further analyzed the output lengths of Qwen3-VL-32B-Thinking and EvoCUA on general benchmarks. The results reveal a significant reduction in EvoCUA's token count compared to Qwen3-VL-32B-Thinking (2,514 vs 3,620), accompanied by a shift in output style.

\textbf{Conclusion.} 
The consistent performance on the OpenCUA backbone validates the effectiveness of our training strategy. The performance decline observed in the Qwen3-VL-Thinking-based variants is primarily attributed to a shift in general data distribution and patterns. Future updates of the EvoCUA models will incorporate an upgraded thinking-based general dataset. This alignment is expected to resolve the current discrepancy and further improve the model generalization performance.

\begin{table*}[t]
\centering
\caption{Performance comparison on the offline grounding and general benchmarks.  Values marked with * are sourced from other public reports.}
\label{tab:combined_results}
\resizebox{\textwidth}{!}{%
\begin{tabular}{l|ccc|ccccc}
\toprule
\multirow{2}{*}{\textbf{Model}} & \multicolumn{3}{c|}{\textbf{GUI Grounding}} & \multicolumn{5}{c}{\textbf{General Multimodal Capabilities}} \\
\cmidrule(lr){2-4} \cmidrule(lr){5-9}
 & \textbf{ScreenSpot v2} & \textbf{ScreenSpot Pro} & \textbf{OSWorld-G} & \textbf{MMMU} & \textbf{MMMU-Pro} & \textbf{MathVista} & \textbf{MMStar} & \textbf{OCRBench} \\
\midrule
OpenCUA-72B & 92.90* & 60.80* & 66.95 & 60.67 & 43.04 & 70.90 & 66.47 & 83.8 \\
Qwen3-VL-8B-Thinking & 90.09 & 46.40* & 56.70* & 74.10* & 60.40* & 81.40* & 75.30* & 81.9* \\
Qwen3-VL-32B-Thinking & 91.11 & 57.10* & 64.00* & 78.10* & 68.10* & 85.90* & 79.40* & 85.5* \\
\midrule
\textbf{EvoCUA-OpenCUA-72B} & 93.47 & 63.24 & 67.65 & 59.22 & 46.51 & 69.40 & 67.80 & 84.05 \\
\textbf{EvoCUA-8B} & 85.21 & 45.39 & 55.08 & 62.11 & 53.30 & 75.80 & 69.07 & 80.30 \\
\textbf{EvoCUA-32B} & 90.40 & 49.76 & 63.86 & 68.11 & 59.16 & 80.40 & 73.20 & 85.35 \\
\bottomrule
\end{tabular}%
}
\end{table*}

\subsection{Ablation Study}
\label{sec:ablation}

To rigorously verify the contribution of each component within the EvoCUA, we conducted extensive ablation studies. We utilized two distinct foundation models, Qwen3-VL-32B-Thinking and OpenCUA-72B, to demonstrate both the efficacy of our specific modules and the universality of the Evolving Experience Learning paradigm.

\subsubsection{Component Analysis on EvoCUA-32B}
We adopt Qwen3-VL-32B-Thinking as our base checkpoint to dissect the cumulative gains from the Unified Action Space, Cold Start, Rejection Fine-Tuning (RFT), and RL. As shown in Table \ref{tab:ablation_qwen}, each stage of the evolutionary cycle yields significant monotonic improvements.

\begin{table}[h]
\centering
\caption{Detailed ablation study on \textbf{EvoCUA-32B}. We show the absolute gain relative to the previous stage.}
\label{tab:ablation_qwen}
\begin{tabular}{lc}
\toprule
\textbf{Stage} & \textbf{Improvement ($\Delta$)} \\
\midrule
\textbf{+ Unified Action Space} & \textcolor{teal}{\textbf{+4.84\%}} \\
\textbf{+ Cold Start} & \textcolor{teal}{\textbf{+2.62\%}} \\
\textbf{+ RFT} & \textcolor{teal}{\textbf{+3.13\%}} \\
\textbf{+ Offline DPO} & \textcolor{teal}{\textbf{+3.21\%}} \\
\textbf{+ Iterative Training} & \textcolor{teal}{\textbf{+1.90\%}} \\
\bottomrule
\end{tabular}
\end{table}

\paragraph{Impact of Action Space \& Cold Start.}
We first quantified the impact of the unified action space through a controlled univariate experiment, comparing the standard SFT baseline against an SFT variant incorporating our refined action definitions. The explicit formulation of the unified action space provides a foundational gain of \textbf{+4.84\%}. By further injecting behavioral priors through cold start training on synthesized high-quality traces, we observe an additional gain of \textbf{+2.62\%}. This validates that grounding the native model with a structured action schema and coherent reasoning patterns is a prerequisite for effective large-scale experience learning.

\paragraph{Efficacy of Evolutionary Learning (RFT \& DPO).}
Transitioning to the active learning phase, Rejection Fine-Tuning (RFT) significantly boosts performance by +3.13\% by consolidating successful experiences. Subsequently, by explicitly addressing failure modes via DPO, we achieve a substantial +3.21\% improvement, highlighting that learning \textit{what not to do} is as critical as learning successful routines. Crucially, performing an additional iteration of the entire evolutionary cycle (stacking another round of RFT and DPO) yields a further +1.90\%. This continuous gain confirms the self-sustaining nature of our paradigm, where the model iteratively refines its capability boundary through recursive synthesis and correction.

\subsubsection{Generalizability on OpenCUA-72B}
To verify the universality of our approach, we applied the same paradigm to the larger OpenCUA-72B model. As detailed in Table \ref{tab:ablation_opencua}, the Evolving Experience Learning paradigm delivers consistent gains across model scales.

\begin{table}[h]
\centering
\caption{Ablation results on \textbf{OpenCUA-72B}, highlighting the robustness of the paradigm across different model scales.}
\label{tab:ablation_opencua}
\begin{tabular}{lc}
\toprule
\textbf{Stage} & \textbf{Improvement ($\Delta$)} \\
\midrule
\textbf{+Cold Start} & \textcolor{teal}{\textbf{+2.14\%}} \\
\textbf{+RFT} & \textcolor{teal}{\textbf{+3.69\%}} \\
\textbf{+Offline DPO} & \textcolor{teal}{\textbf{+3.02\%}} \\
\textbf{+Iterative Training} & \textcolor{teal}{\textbf{+1.82\%}} \\
\bottomrule
\end{tabular}
\end{table}

The results on OpenCUA-72B echo our findings on Qwen3-VL, with DPO (+3.02\%) and RFT (+3.69\%) providing strong contributions. Interestingly, we observed that pure RFT (stacking 3 rounds without explicit cold start) achieved a remarkable gain of \textbf{+8.12\%} shown in Table \ref{tab:rft_scaling}. This suggests that with a sufficiently strong base model, the synthesis engine and scalable interaction infrastructure alone can drive massive capability improvements, even without explicit prior injection. In addition, OpenCUA-72B adopts the standard pyautogui format. This action space natively supports stateful operations (such as shift+click) and possesses no obvious functional deficiencies.

\subsection{Scaling Analysis}

We investigate the scalability of EvoCUA by analyzing the performance gain ($\Delta\%$) across varying Pass@\textit{k} values, max inference steps and data volume.

\begin{figure}[htbp]
    \centering
    
    \begin{subfigure}[b]{0.48\textwidth}
        \centering
        \begin{tikzpicture}
            \begin{axis}[
                width=\linewidth, height=6.5cm,
                xmin=0.5, xmax=5.5,
                ymin=0, ymax=6.8,
                axis lines=left,
                axis line style={black, thick, -stealth},
                ymajorgrids=true,
                grid style={dashed, gray!20},
                xlabel={\textbf{Pass@\textit{k}} ($k$)},
                ylabel={\textbf{Performance Gain} ($\Delta\%$)},
                label style={font=\small},
                ylabel style={yshift=0pt},
                xtick={1, 2, 3, 4, 5},
                xticklabels={8, 16, 32, 64, 128},
                legend style={
                    at={(0.98,0.98)}, 
                    anchor=north east, 
                    draw=none, 
                    fill=white!90, 
                    font=\tiny, 
                    legend cell align=left 
                }
            ]

            \addplot[color=blue, thick, mark=*, mark options={scale=1.2, fill=blue, draw=white, line width=1pt}] 
            coordinates { (1, 4.55) (2, 4.93) (3, 2.39) (4, 2.99) (5, 2.35) };
            \addlegendentry{EvoCUA-32B - Qwen3-VL-32B-Thinking}

            \addplot[color=blue, thick, dashed, mark=x, mark options={scale=1.5, thick, solid}] 
            coordinates { (1, 4.21) (2, 4.65) (3, 4.37) (4, 3.74) (5, 3.43)};
            \addlegendentry{EvoCUA-8B - Qwen3-VL-8B-Thinking}

            \node[above, font=\tiny] at (axis cs:1, 4.55) {+4.55\%};
            \node[above, font=\tiny] at (axis cs:2, 4.93) {+4.93\%};
            \node[above, font=\tiny] at (axis cs:3, 2.39) {+2.39\%};
            \node[below, font=\tiny] at (axis cs:4, 2.99) {+2.99\%};
            \node[above, font=\tiny] at (axis cs:5, 2.35) {+2.35\%};

            \node[below, font=\tiny, yshift=-2pt] at (axis cs:1, 4.21) {+4.21\%};
            \node[below, font=\tiny, yshift=-2pt] at (axis cs:2, 4.65) {+4.65\%};
            \node[above, font=\tiny] at (axis cs:3, 4.37) {+4.37\%};
            \node[above, font=\tiny] at (axis cs:4, 3.74) {+3.74\%};
             \node[above, font=\tiny] at (axis cs:5, 3.43) {+3.43\%};

            \end{axis}
        \end{tikzpicture}
        \caption{Performance Gain across Pass@\textit{k}. The Y-axis displays the absolute gain of EvoCUA over the Qwen3-VL-Thinking baseline.}
        \label{fig:pass_k}
    \end{subfigure}
    \hfill 
    \begin{subfigure}[b]{0.48\textwidth}
        \centering
        \begin{tikzpicture}
            \begin{axis}[
                width=\linewidth, height=6.5cm,
                xmin=0.8, xmax=4.2,
                ymin=0, ymax=20,
                axis lines=left,
                axis line style={black, thick, -stealth},
                ymajorgrids=true,
                grid style={dashed, gray!20},
                xlabel={\textbf{Max Inference Steps}},
                ylabel={\textbf{Performance Gain} ($\Delta\%$)},
                label style={font=\small},
                ylabel style={yshift=0pt},
                xtick={1, 2, 3, 4},
                xticklabels={15, 30, 50, 100},
                legend style={
                    at={(0.98,0.05)}, 
                    anchor=south east, 
                    draw=none, 
                    fill=white!90, 
                    font=\tiny,
                    legend cell align=left
                },
                visualization depends on={value \coordindex \as \myindex},
                nodes near coords={%
                    \ifnum\myindex=0%
                    \else%
                       +\pgfmathprintnumber\pgfplotspointmeta\%
                    \fi%
                },
                nodes near coords style={font=\tiny, color=black}
            ]

            \addplot[color=blue, thick, mark=*, mark options={scale=1.2, fill=blue, draw=white, line width=1pt}, nodes near coords align={above}] 
            coordinates { (1, 0.00) (2, 10.71) (3, 16.25) (4, 17.36) };
            \addlegendentry{EvoCUA-32B - EvoCUA-32B (Step 15)}

            \addplot[color=blue, thick, dashed, mark=x, mark options={scale=1.5, thick, solid}, nodes near coords align={below}, every node near coord/.append style={yshift=-2pt}] 
            coordinates { (1, 0.00) (2, 7.58) (3, 10.25) (4, 10.84) };
            \addlegendentry{EvoCUA-8B - EvoCUA-8B (Step 15)}

            \node[anchor=south west, font=\tiny, gray, xshift=-5pt] at (axis cs:1, 0.8) {Baseline};

            \end{axis}
        \end{tikzpicture}
        \caption{Scaling with Inference Steps. The Y-axis represents the absolute gain relative to the performance at \textit{step=15}.}
        \label{fig:inference_steps}
    \end{subfigure}

    \caption{\textbf{Performance analysis of EvoCUA models.} (a) Improvement over the base model across varying Pass@\textit{k} metrics. Legends indicate the specific backbone models used. (b) Performance scaling with increased maximum inference steps. The legends denote the performance gain relative to the Step 15 baseline. The 32B model shows significantly stronger scaling capabilities.}
    \label{fig:combined_analysis}

\end{figure}
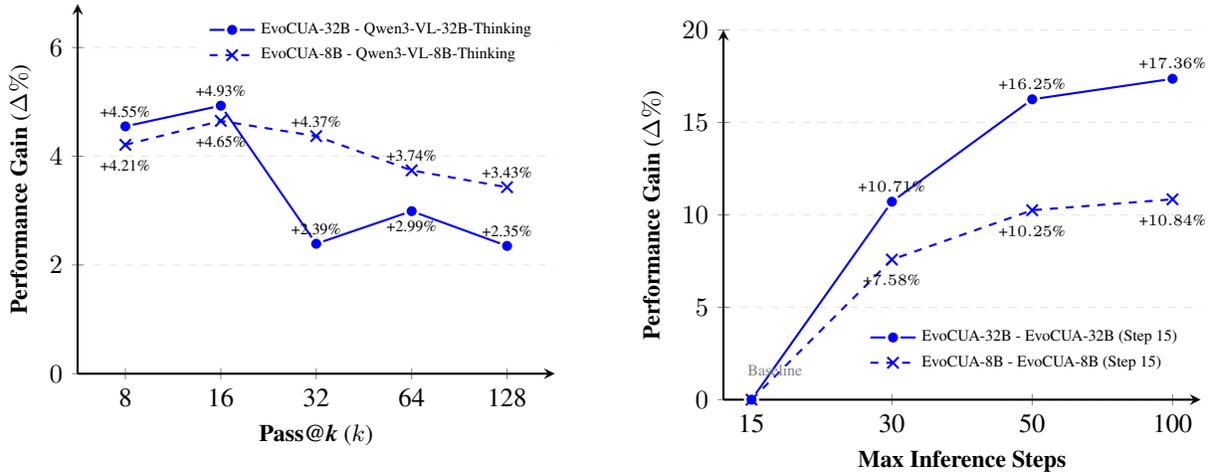

\textbf{Scaling with Pass@\textit{k}}. In figure \ref{fig:pass_k}, EvoCUA maintains a consistent performance lead over the base model (Qwen3-VL-Thinking) across all Pass@\textit{k} metrics. As depicted in Figure \ref{fig:pass_k}, the 32B model sustains a positive gain, peaking at +4.93\% at $k=16$ and maintaining a significant advantage even at higher $k$ values. This consistent gap demonstrates that our training strategy optimizing the action space and reasoning priors fundamentally elevates the model's performance ceiling.

\textbf{Scaling with Max Steps}. In figure \ref{fig:inference_steps}, We observe that performance steadily improves as the maximum step limit increases. Increasing the inference capacity from 15 to 50 steps leads to consistent gains, with the 32B model achieving a +16.25\% improvement over the baseline. Beyond 50 steps, the rate of improvement moderates, primarily due to the scarcity of trajectories exceeding 50 steps in the current training distribution.

\textbf{Experience Scaling}. We conduct experience scaling experiments on RFT. Specifically, we perform an ablation study on an early iteration of the OpenCUA-72B model, omitting the cold-start and dpo phase to focus exclusively on multi-round RFT. As shown in Table \ref{tab:rft_scaling}, the performance gains relative to the baseline are as follows:
\begin{itemize}

\item \textit{Round 1}: Independent training on 20k samples yields a \textbf{+2.61 pp} gain.

\item \textit{Round 2}: Iterative training on 226k samples, initialized from Round 1 checkpoint, increases the gain to \textbf{+6.79 pp}.

\item \textit{Round 3}: Training the OpenCUA-72B base on 1M samples aggregated from three RFT iterations achieves an \textbf{+8.12 pp} improvement.

\end{itemize}

Our analysis highlights a critical trade-off between data scale, off-policy distribution, and the signal-to-noise ratio (SNR). As model capabilities improve with scale, the tolerance for noise decreases, creating a bottleneck for existing iterative methods. Crucially, however, we remain confident that further scaling can be sustained, provided that data quality, on-policy alignment, and SNR are effectively optimized.

\textbf{Environmental Uncertainty and Evaluation}. It is critical to distinguish the role of Pass@\textit{k} in agentic tasks versus standard LLM benchmarks. In traditional text generation, the "environment" (the prompt) is static and deterministic; thus, Pass@\textit{k} solely measures the diversity of the model's internal capacity. In contrast, GUI environments introduce inherent environmental stochasticity. Factors such as system latency, network fluctuations, and minor rendering variations mean that identical action sequences can yield different state transitions. Consequently, in this context, Pass@\textit{k} serves a dual purpose: it evaluates not only the model's generative diversity but also its robustness against environmental noise. We observe that even with deterministic sampling (temperature=0), success rates exhibit variance due to these system perturbations. This finding highlights a critical limitation of pure data scaling. To achieve human-level reliability, future research must prioritize environment scaling—expanding environmental diversity and modeling dynamic uncertainties to ensure robustness across real-world systems.

\begin{table}[h]
    \centering
    \caption{Experience scaling results on RFT. The absolute gains are all relative to the baseline.}
    \label{tab:rft_scaling}
    
    \begin{tabular}{l c l r}
        \toprule
        \textbf{Stage} & \textbf{Data Size} & \textbf{Gain ($\Delta$\%)} \\
        \midrule
        RFT Round 1 & 20k & +2.61 \\
        RFT Round 2 & 226k  & +6.79 \\
        RFT Round 3 & 1M & +8.12 \\
        \bottomrule
    \end{tabular}
\end{table}

\subsection{Discussions}
\label{sec:insights}

Drawing from over \textbf{a thousand individual experiments totaling more than 1 million accelerator hours}, we categorize our observations regarding the training dynamics of native computer use agents into four critical dimensions.

\paragraph{1. The Dual Nature of Experiences.}
Our analysis reveals that the signal-to-noise ratio varies fundamentally between success and failure trajectories, necessitating distinct processing strategies.
\begin{itemize}
    \item \textit{Success trajectories:} Trajectories generated by the model represent known knowledge characterized by low noise but limited information gain. While the final outcome is correct, step-level redundancy constitutes a major noise source. Without aggressive filtering of these inefficient steps, the model becomes fragile, leading to phenomena such as action aliasing (outputting conflicting actions for a single state) and cyclic repetition (endlessly clicking the same coordinates). Effective filtering is thus a prerequisite for multi-round rejection sampling fine-tuning.
    \item \textit{Failure trajectories:} Conversely, failure trajectories are high-noise but high-information. They delineate the model's capability boundaries and contain corner cases that the current policy cannot handle. While raw failure data is too noisy for direct learning, identifying critical error steps allows for the construction of preference pairs. This transforms failed attempts into a high-value source for boundary alignment.
\end{itemize}

\paragraph{2. Foundational Constraints and Initialization.}
The initialization phase substantially influences the agent's potential performance.
\begin{itemize}
    \item \textit{Completeness of action space:} A comprehensive definition of the action space is a prerequisite. Missing high-efficiency operations (e.g., triple click, shift-based shortcuts) renders specific tasks, such as complex spreadsheet editing, effectively unsolvable. Post-hoc additions to the action space are inefficient compared to a correct initial definition.
    \item \textit{Pattern-centric cold start:} The cold start phase should prioritize pattern diversity over data volume. We observed that a lightweight cold start is sufficient to establish a latent alignment—grounding the action space and stabilizing output formatting. A heavy cold start often yields high supervised metrics but creates a checkpoint that is harder to refine later. A lightweight initialization, followed by rigorous rejection sampling and preference optimization, consistently produces superior final performance.
\end{itemize}

\paragraph{3. Dynamics of Iterative Optimization.}
Computer use tasks are inherently long-horizon, often requiring dozens of interaction turns. Optimizing for this requires strict adherence to specific dynamic properties.
\begin{itemize}
    \item \textit{The on-policy imperative:} We emphasize the necessity of using strictly on-policy data during iterative learning. We hypothesize that off-policy data disrupts the principal direction of the optimization vector established during supervision. Once the model's weights diverge from the optimal manifold due to distribution shifts, recovering the correct optimization path is computationally prohibitive.
    \item \textit{Termination asymmetry:} The distribution of the terminate action is the most critical control variable. We observed a distinct asymmetry: the model converges rapidly on failure recognition, whereas recognizing success requires a carefully calibrated density of positive samples. An excessive concentration of success signals leads to premature termination, while a deficit prevents the agent from stopping.
    \item \textit{Self-correction and future potential:} To mitigate error accumulation in long-horizon tasks, we utilize preference optimization focused on state checking and reflection. By targeting steps where the agent fails to perceive errors, we enhance robustness. These improvements suggest that the logical evolution is a transition to online reinforcement learning, where advanced credit assignment mechanisms can further optimize performance in complex, multi-step environments.
\end{itemize}

\paragraph{4. Visualization-Driven Diagnosis and Iteration.} We argue that achieving SOTA performance in long-horizon tasks requires more than algorithmic novelty; it demands a transparent debugging infrastructure. We developed a comprehensive suite of trajectory analysis and visualization tools that served as the "eyes" of our evolutionary cycle. These tools played a pivotal role in three critical phases:

\begin{itemize}
\item \textit{Quality Assurance for Synthesis}: They allowed us to visualize synthesized samples alongside their ground-truth states, enabling rapid identification of "hallucinated validators" or executable logic errors in our Synthesis Engine before they polluted the training pool.

\item \textit{Cold-Start Data Construction}: By visually contrasting the trajectory characteristics of different foundation models, we identified superior reasoning patterns and action sequences. This guided the curation of our high-quality Cold Start dataset, ensuring the agent learned robust behavioral priors rather than noisy imitation.

\item \textit{Failure Analysis for Refinement}: Our Pass@k Differential Analysis tool aggregates successful and failed trajectories for the same query. This granular comparison helped us pinpoint specific failure modes—such as coordinate drift or reasoning-action misalignment—directly informing the design of our step-level policy optimization to rectify these specific weaknesses.
\end{itemize}

%% file: sections/onlinerl.tex
\section{Future Work on Online Agentic RL}
Reinforcement Learning with Verifiable Rewards (RLVR)~\citep{guo2025deepseek} has become a crucial framework for boosting the reliability, generalization, and performance of model. Building on this, our future work aims to explore online agentic reinforcement learning in GUI-based agent tasks. Constrained by time limitations, we have not yet conducted sufficient model training and comprehensive benchmark evaluations. Accordingly, the subsequent parts of this section will first conduct an in-depth analysis of the training-inference discrepancy issue, and then discuss the future research directions to advance this work.

\paragraph{Training-Inference Discrepancy in Trajectory-Level Training}

Algorithms such as GRPO~\citep{shao2024deepseekmath} have been shown to be effective on a wide range of reasoning tasks. These algorithms collect a set of trajectories for a single query, calculate the advantage function within the trajectory group, and conduct training at the trajectory granularity. However, trajectory-level training will cause training-inference discrepancy in GUI tasks. During the rollout phase, GUI model does not retain all complete context information, but only preserves the complete information of recent steps (including screenshots, reasoning and actions), while earlier historical information is compressed into text-only semantic actions. If the trajectory of the final step is directly used for training, the model will not be able to learn the supervision signals of intermediate steps.

\paragraph{Step-Level Policy Optimization}
To address the training-inference discrepancy in trajectory-level training, we propose namely \textbf{Ste}p-Level \textbf{P}olicy \textbf{O}ptimization (\textbf{STEPO}), a simple yet effective policy optimization algorithm. 

For a trajectory $\tau$ with length $T$, each step $t\in\{1,2,\dots,T\}$ contains $K_t$ tokens. We denote the $k$-th token in step $t$ as $x_{t,k}$ ($k\in\{1,2,...,K_t\}$), and the full token sequence of step $t$ is represented by $x_t=(x_{t,1},x_{t,2},\dots,x_{t, K_t})$. For the trajectory set $\mathcal{T}=\{\tau_1, \tau_2, \dots, \tau_n\}$, the token at position $k$ of step $t$ in the $i$-th trajectory is denoted as $x_{i,t,k}$. 

For each question $q$, similar to GRPO, STEPO samples a group $G$ of trajectories $\{\tau_1, \tau_2, \dots, \tau_n\}$ and calculates the advantages within the trajectory group:
\begin{equation}
\hat{A_i}=\frac{R_i-\text{mean}(\{R_j\}_{j=1}^G)}{\text{std}(\{R_j\}_{j=1}^G)}
\end{equation}
Where $R_i$ represents the reward of the trajectory $\tau_i$. Subsequently, the advantage value $\hat{A}_i$ corresponding to each trajectory $\tau_i$ is uniformly allocated to all steps contained in the trajectory, that is:
\begin{equation}
    \hat{A}_{i,t}=\hat{A}_i/T_i, t \in \{1, 2,\dots, T_i\},
\end{equation}
where $T_i$ denotes the number of steps contained in the trajectory $\tau_i$. All tokens within the same step share the corresponding advantage value $A_{i,t}$ of this step. On this basis, we conduct model training using all step-level samples. The optimization objective of the proposed algorithm can be demonstrated as:
\begin{align}
\tiny
\mathcal{J}&_{\text{STEPO}}(\theta)=\mathbb{E}[q\sim P(Q),\notag\{\tau_i\}_{i=1}^G\sim\pi_{\theta_{old}}(\mathcal{T}|q)]\\
&\frac{1}{G}\sum_{i=1}^G\sum_{t=1}^{T_i}\frac{1}{K_t}\sum_{k=1}^{K_t}\{\min[r_{i,t,k}(\theta)\hat{A}_{i,t}, \text{clip}(r_{i,t,k}, 1-\epsilon_{\text{low}}, 1+\epsilon_{\text{high}})\hat{A}_{i, t}]-\beta\mathbb{D}_{KL}(\pi_\theta\|\pi_{\text{ref}})\},
\end{align}
where
\begin{align}
r_{i,t,k}(\theta)=\frac{\pi_{\theta}(\tau_{i,t,k}|q, \tau_{i,t,<k})}{\pi_{\theta_{\text{old}}}(\tau_{i,t,k}|q, \tau_{i,t,<k})},
\end{align}
denotes the importance sampling ratio. $\epsilon$ denotes the clipping parameter,  $\mathbb{D}_{KL}$ denotes a KL penalty term and $\beta$ controls the KL divergence regularization. By uniformly allocating the advantage value of a trajectory to all steps it comprises, this strategy achieves two core optimization effects: first, it drives high-advantage-value trajectories to complete tasks with fewer steps, thereby reducing redundant execution steps; second, it prompts low-advantage-value trajectories to expand the number of exploration steps, so as to improve the task completion rate. By the step-level policy optimization mechanism, STEPO can effectively circumvent the training-inference discrepancy issue.

\begin{figure}[htbp]
    \centering
    \includegraphics[width=0.65\linewidth]{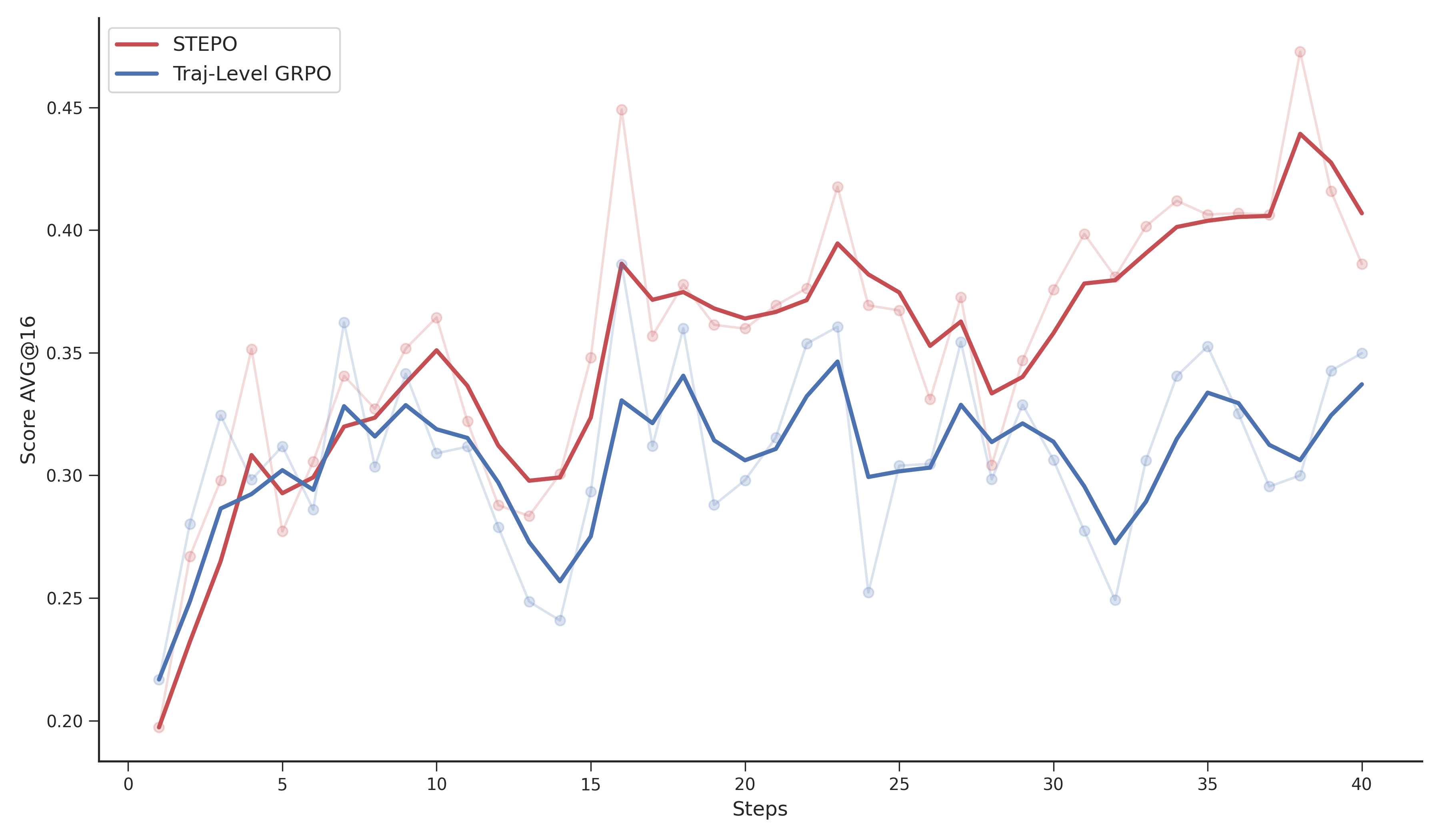}
    \caption{The values illustrated in this figure denote the 16-time average scores of the two methods, respectively.}
    \label{fig:online_rl}
\end{figure}

\paragraph{Experiments and Analysis}
To clarify the impact of train-inference discrepancy and verify the effectiveness of STEPO, we conduct online RL training on the OpenCUA-32B model. As illustrated in the figure~\ref{fig:online_rl}, the training performance of STEPO is significantly superior to that of GRPO trained with final trajectories, which fully confirms the effectiveness of STEPO.

However, STEPO suffers from the issue of high training cost, as the number of updates to the policy model multiplies significantly. Accordingly, we hypothesize that the requirements for step-level training may not be uniform across different training phases, and training only specific key steps might also achieve comparable performance to training all steps. In the future, we will explore directions such as scaling up online RL and developing more effective RL training recipes.

%% file: sections/related_work.tex
\section{Related Work}
\textbf{Foundation VLMs and Computer Use Capabilities}. The landscape of Large Visual Language Models (VLMs) has rapidly evolved to support complex agentic tasks. Proprietary frontier models, most notably Claude 4.5 Sonnet~\citep{anthropic2025claude45} and Seed 1.8~\citep{bytedance2025seed18}, have set the industry standard, demonstrating human-level proficiency in zero-shot instruction following and long-horizon planning. In the open-weight domain, Qwen3-VL~\citep{Qwen3-VL} has emerged as a robust backbone, introducing next-generation dynamic resolution and enhanced OCR capabilities. EvoCUA builds directly on the Qwen3-VL architecture, enhancing it via a specialized evolutionary post-training curriculum to transcend general-purpose pre-training limitations.

\textbf{Generalist GUI Agents and Benchmarks}. To evaluate online agent performance, OSWorld~\citep{xie2024osworld} serve as primary testbeds. OpenCUA~\citep{wang2025opencua} establishes a critical foundation with the AgentNet dataset, while state-of-the-art efforts like UI-TARS-2~\citep{wang2025ui} and Step-GUI~\citep{yan2025step} utilize multi-turn RL and step-wise visual reasoning, respectively. Unlike these demonstration-heavy approaches, EvoCUA utilizes autonomously synthesized, verifiable experiences to reduce annotation costs while achieving superior performance on the OSWorld leaderboard.

\textbf{Visual Grounding and Action Execution}. Precise GUI grounding remains a cornerstone of native computer use. Early approaches like Aguvis~\citep{xu2024aguvis} laid the groundwork, while recent models such as ShowUI~\citep{lin2025showui} and UGround~\citep{gou2024navigating} have optimized vision-language-action architectures specifically for high-resolution layouts. EvoCUA incorporates insights from these grounding-specialized architectures to establish robust execution primitives prior to high-level planning optimization.

\textbf{From Imitation to Learning from Experience}. Training paradigms are shifting from Behavior Cloning (BC) toward Reinforcement Learning (RL). While standard algorithms like PPO~\citep{schulman2017proximal} have been successfully adapted for multi-turn GUI interaction by UI-TARS-2~\citep{wang2025ui}, recent research focuses on incentivizing reasoning capabilities. This transition was pioneered by DeepSeek-R1~\citep{guo2025deepseek} and DeepSeekMath~\citep{shao2024deepseekmath}, which introduced the Reinforcement Learning with Verifiable Rewards (RLVR) paradigm. They demonstrated that RL can implicitly verify complex reasoning chains without dense process supervision. Following this, Feng et al.~\citep{feng2025group} proposed Group-in-Group optimization to stabilize such training, while Zhang et al.~\citep{zhang2025agent} explored learning via reward-free "Early Experience."  EvoCUA advances this direction by addressing the data scarcity bottleneck through a verifiable synthesis engine, which autonomously produces scalable, ground-truth-verified synthetic data. This foundation enables our evolving paradigm via learning from experience, a self-sustaining cycle that iteratively enhances agent capabilities through large-scale rejection sampling and preference learning on verifiable synthetic trajectories.

%% file: sections/conclusion.tex
\section{Conclusion}

In this work, we present EvoCUA, a native computer use agent developed through the evolving paradigm via learning from experience. By integrating verifiable synthesis with a scalable interaction infrastructure, we demonstrate the efficacy of converting synthetic compute into high-quality training signals. Empirical evaluations on the OSWorld benchmark validate this approach, with EvoCUA achieving a success rate of 56.7\%, establishing a new state-of-the-art among open-weights models.

Despite these advancements, a performance gap persists between current open models and leading closed-weights systems or human-level reliability. This disparity highlights the limits of offline learning from synthesized traces alone. To address this, our preliminary investigation into online reinforcement learning identifies active environmental interaction as a critical driver for further improvement, evidenced by a consistent upward trend in reward accumulation. Future work will focus on systematically expanding this online evolutionary boundary, aiming to bridge the remaining gap and achieve fully autonomous computer use capabilities.

%% file: sections/acknowledgements.tex
\newpage
\section*{Acknowledgments}

We sincerely thank the open-source community for their significant contributions to the computer use agent field. 
We are particularly grateful to Xinyuan Wang and Tianbao Xie, the core authors of OpenCUA and OSWorld respectively, for their insightful discussions, valuable feedback on evaluation, and continuous support throughout this project. Their pioneering work has greatly inspired and advanced our research. We are committed to giving back to the community and will continue to open-source our research to advance the field.

We also thank our colleagues and family members listed below. We truly appreciate their constant support, encouragement, and helpful discussions throughout this project. The listing is in alphabetical order by first name:

\vspace{0.5em} 

\begin{multicols}{2}
\noindent
Chen Gao \\
Daorun Pan \\
Jiahui Wang \\
Jiangke Fan \\
Jiarong Shi \\
Kefeng Zhang \\
Rumei Li \\
Wenlong Zhu \\
Xuejia Shi \\
Xuezhi Cao \\
Ying Ouyang \\
Yerui Sun \\
Yuchao Zhu \\
Yufei Zhang \\
Yuwei Jiang
\end{multicols}

\vspace{0.5em}

%% file: sections/appendix.tex
\section{Unified Action Space}
\label{appendix:action_space}

The following table details the unified native action space $\mathcal{A}$ implemented in EvoCUA. The agent interacts with the environment by invoking the \texttt{computer\_use} function with a specific \texttt{action} and its corresponding arguments.

\begin{table}[h]
\centering
\caption{Detailed Definition of the EvoCUA Native Action Space}
\label{tab:action_space}
\resizebox{\textwidth}{!}{%
\begin{tabular}{@{}llp{8cm}l@{}}
\toprule
\textbf{Category} & \textbf{Action Primitive} & \textbf{Description} & \textbf{Required Arguments} \\ \midrule
\multirow{4}{*}{\textbf{Keyboard}} & \texttt{key} & Performs a key press and release sequence on the specified keys. & \texttt{keys} (array) \\
 & \texttt{key\_down} & Presses and \textbf{holds} the specified key(s). Used for stateful operations (e.g., holding Shift). & \texttt{keys} (array) \\
 & \texttt{key\_up} & Releases the specified key(s) in reverse order. & \texttt{keys} (array) \\
 & \texttt{type} & Types a string of text on the keyboard. & \texttt{text} (string) \\ \midrule
\multirow{8}{*}{\textbf{Mouse}} & \texttt{mouse\_move} & Moves the cursor to the specified pixel coordinates. & \texttt{coordinate} (x, y) \\
 & \texttt{left\_click} & Clicks the left mouse button at the specified coordinates. & \texttt{coordinate} (x, y) \\
 & \texttt{right\_click} & Clicks the right mouse button at the specified coordinates. & \texttt{coordinate} (x, y) \\
 & \texttt{middle\_click} & Clicks the middle mouse button at the specified coordinates. & \texttt{coordinate} (x, y) \\
 & \texttt{double\_click} & Double-clicks the left mouse button at the specified coordinates. & \texttt{coordinate} (x, y) \\
 & \texttt{triple\_click} & Triple-clicks the left mouse button (useful for text selection). & \texttt{coordinate} (x, y) \\
 & \texttt{left\_click\_drag} & Clicks and drags the cursor to the target coordinates. & \texttt{coordinate} (x, y) \\
 & \texttt{scroll} / \texttt{hscroll} & Performs a vertical or horizontal scroll. & \texttt{pixels} (number) \\ \midrule
\multirow{3}{*}{\textbf{Control}} & \texttt{wait} & Pauses execution for a specified duration to allow UI rendering. & \texttt{time} (number) \\
 & \texttt{terminate} & Terminates the current task and reports the final status. & \texttt{status} ("success"|"failure") \\ \bottomrule
\end{tabular}%
}
\end{table}

\section{Cold Start: Hindsight Reasoning Generation}
\label{sec:hindsight_reasoning}

To construct high-quality data for the supervised cold-start phase, we transform raw physical interaction traces into training samples augmented with explicit cognitive chains. We employ a \textbf{Hindsight Reasoning Generation} strategy to achieve this. By treating the ground-truth execution path as known future information, we utilize a general model to retrospectively generate reasoning traces ($z_t$) that explain the observed actions, thereby establishing a causal alignment between cognition and execution.

The generation process is driven by a series of context-aware prompt templates that enforce the structural schemas defined in our \textbf{Thought Space ($\mathcal{Z}$)}. Depending on the execution phase, the generation logic adapts as follows:

\paragraph{1. Goal Clarification ($z_0$)}
At the initial step of a trajectory ($t=0$), the reasoning generation focuses on resolving ambiguity and establishing a global plan.
\begin{itemize}
    \item \textbf{Context:} The general model is provided with the user instruction, the initial screenshot, and the first executable code block.
    \item \textbf{Generation Logic:} We utilize a specific template that enforces a first-person perspective. The model must explicitly state the current environment state, clarify the task goal, and articulate a high-level plan (e.g., ``\textit{I need to open the browser to search for...}'') before justifying the specific action taken. This ensures that the subsequent physical execution is grounded in a clear intent.
\end{itemize}

\paragraph{2. Observation Consistency ($z_{obs}$)}
For intermediate steps, the objective is to maintain semantic consistency between the visual observation and the reasoning trace.
\begin{itemize}
    \item \textbf{Context:} The model analyzes the transition from the previous state to the current state.
    \item \textbf{Generation Logic:} The prompt instructs the model to identify \textit{``What changed''} in the environment and explain \textit{``Why this action is needed''} to advance the workflow.
    \item \textbf{Semantic Abstraction:} To prevent overfitting to specific screen resolutions, the prompt explicitly constrains the generation to avoid mentioning raw pixel coordinates. Instead, the model is guided to describe target UI elements semantically (e.g., ``\textit{Click on the `File' menu}'' rather than ``\textit{Click at (100, 200)}''), ensuring the reasoning remains robust to layout variations.
\end{itemize}

\paragraph{3. Reflection and Correction ($z_{reflect}$)}
For trajectories involving error recovery (``resume'' traces), we implement a specialized \textbf{Reflection Mechanism}.
\begin{itemize}
    \item \textbf{Context:} When processing a trajectory segment that recovers from a failure, the synthesis engine injects the specific \texttt{analysis\_reason} (the root cause of the prior failure) into the prompt context.
    \item \textbf{Generation Logic:} The model is enforced to begin the thought trace with a dedicated header: ``\textbf{Reflection: }''. It must retrospectively analyze the failure (e.g., ``\textit{Reflection: I realize that my previous attempt to click the icon failed because...}'').
    \item \textbf{Self-Correction:} Following the reflection, the model must naturally transition to a corrected plan (e.g., ``\textit{Now I will try a different approach...}''), effectively internalizing the logic of self-correction into the training data.
\end{itemize}

\paragraph{4. Reasoning-Augmented Termination ($z_T$)}
To mitigate premature or delayed stopping, the termination action is conditioned on a rigorous visual verification process.
\begin{itemize}
    \item \textbf{Context:} The generation is triggered at the final step of a trajectory.
    \item \textbf{Generation Logic:} The general model is required to assess the final screenshot against the initial instruction. It must generate a reasoning trace that provides visual evidence of task completion (or failure) before emitting the final \texttt{terminate} signal. This ensures that the agent's termination decision is grounded in logical verification rather than memorized trajectory lengths.
\end{itemize}

\begin{algorithm}[h]
\caption{Hindsight Reasoning Generation}
\label{alg:hindsight_reasoning}
\begin{algorithmic}[1]
\Require 
    Instruction $g$; 
    Raw Trajectory $\tau = \{(o_t, a_t)\}_{t=0}^T$; 
    Error Context $c_{err}$ (optional, for resume traces)
\Ensure 
    Reasoning Traces $\mathcal{Z} = \{z_t\}_{t=0}^T$

\State $\mathcal{Z} \leftarrow \emptyset$
\State $h_{prev} \leftarrow \emptyset$ \Comment{Initialize interaction history}

\For{$t \leftarrow 0$ \textbf{to} $T$}
    \State $prompt \leftarrow \text{NULL}$
    
    \If{$t = 0$} \Comment{\textbf{Phase 1: Initialization}}
        \If{$c_{err} \neq \text{NULL}$}
            \State \Comment{Trigger Reflection Mechanism for error recovery}
            \State $prompt \leftarrow \textsc{ConstructReflectPrompt}(g, c_{err}, o_0, a_0)$
        \Else
            \State \Comment{Standard Goal Clarification}
            \State $prompt \leftarrow \textsc{ConstructGoalPrompt}(g, o_0, a_0)$
        \EndIf
        
    \ElsIf{$t = T$} \Comment{\textbf{Phase 3: Termination}}
        \State \Comment{Reasoning-Augmented Termination Verification}
        \State $prompt \leftarrow \textsc{ConstructTermPrompt}(g, h_{prev}, o_T)$
        
    \Else \Comment{\textbf{Phase 2: Intermediate Execution}}
        \State \Comment{Ensure Observation Consistency}
        \State $prompt \leftarrow \textsc{ConstructObsPrompt}(g, h_{prev}, o_t, a_t)$
    \EndIf
    
    \State \Comment{Query General Model}
    \State $z_t \leftarrow \text{GeneralLLM}(prompt)$ 
    \State $\mathcal{Z} \leftarrow \mathcal{Z} \cup \{z_t\}$
    \State $h_{prev} \leftarrow h_{prev} \cup \{(z_t, a_t)\}$
\EndFor

\State \textbf{return} $\mathcal{Z}$
\end{algorithmic}
\end{algorithm}

\section{Algorithm for DPO}
In this section, we present the algorithmic implementation of Step-Level Direct Preference Optimization (DPO). This method focuses on two core processes: \textbf{Key Error Identification} and \textbf{Preference Pair Construction}. Algorithm \ref{alg:dpo_main} details how we identify Critical Forking Points from failure trajectories and construct paired data for both Action Correction and Reflection.

\begin{algorithm}
\caption{Step-Level DPO Pair Construction}
\label{alg:dpo_main}
\begin{algorithmic}[1]
\Require Target Trajectory $\mathcal{T}_{tgt}$ (Failure Case), Reference Trajectory $\mathcal{T}_{ref}$ (Success Case)
\Require VLM $\mathcal{M}$ (for alignment and synthesis)
\Ensure DPO Dataset $\mathcal{D}_{dpo}$

\State \Comment{\textbf{Step 1: Error Identification}}
\State $E \leftarrow \text{AnalyzeErrorSteps}(\mathcal{T}_{tgt}, \mathcal{T}_{ref})$ 
\State $\mathcal{D}_{dpo} \leftarrow \emptyset$

\For{\textbf{each} error step index $t$ in $E$}
    \State \Comment{Extract context from the failure trajectory}
    \State $o_{t}, a_{rejected} \leftarrow \text{GetStateAndAction}(\mathcal{T}_{tgt}, t)$
    
    \State \Comment{\textbf{Step 2: Critical Forking Point Discovery}}
    \State $S_{aligned} \leftarrow \text{None}$
    \For{$k \leftarrow t-w$ \textbf{to} $t+w$}
        \State $o_{ref}, a_{ref} \leftarrow \text{GetStateAndAction}(\mathcal{T}_{ref}, k)$
        \If{$\text{CheckAlignment}(\mathcal{M}, o_{t}, a_{rejected}, a_{ref})$ \textbf{is True}}
            \State $S_{aligned} \leftarrow \text{NormalizeCoords}(a_{ref}, o_{t})$ 
            \State \textbf{break}
        \EndIf
    \EndFor
    
    \If{$S_{aligned} \neq \text{None}$}
        \State \Comment{\textbf{Step 3: Construct Paradigm I (Correction)}}
        \State $z_{enhanced} \leftarrow \mathcal{M}.\text{SynthesizeThought}(o_{t}, S_{aligned})$
        \State $\tau_{chosen} \leftarrow (z_{enhanced}, S_{aligned})$
        \State $\mathcal{D}_{dpo}.\text{add}(\{ \text{state}: o_{t}, \text{chosen}: \tau_{chosen}, \text{rejected}: a_{rejected} \})$
        
        \State \Comment{\textbf{Step 4: Construct Paradigm II (Reflection)}}
        \If{$t+1 < \text{Length}(\mathcal{T}_{tgt})$}
            \State $o_{next}, a_{blind} \leftarrow \text{GetStateAndAction}(\mathcal{T}_{tgt}, t+1)$
            \State \Comment{Observe Error $\to$ Stop $\to$ Plan}
            \State $z_{reflect} \leftarrow \mathcal{M}.\text{GenerateReflection}(o_{next}, a_{rejected}, S_{aligned})$
            \State $\tau_{chosen\_ref} \leftarrow (z_{reflect}, S_{aligned})$
            \State $\mathcal{D}_{dpo}.\text{add}(\{ \text{state}: o_{next}, \text{chosen}: \tau_{chosen\_ref}, \text{rejected}: a_{blind} \})$
        \EndIf
    \EndIf
\EndFor
\State \Return $\mathcal{D}_{dpo}$
\end{algorithmic}
\end{algorithm}

\section{Trajectory Analysis and Visualization}
\label{appendix:vis_tool}

To enable granular diagnosis of agent behaviors and strictly validate the quality of our \textbf{synthetically generated experience}, we developed the \textbf{EvoCUA Trajectory Inspector}. This visualization system allows us to examine the frame-by-frame alignment between the agent's visual observation ($o_t$), internal reasoning trace ($z_t$), and the executable code action ($a_t$).

We illustrate the utility of this system using a representative synthetic task from the spreadsheet domain: \textit{"Find the greatest value per row and place it in Column G."} This long-horizon task serves as a rigorous testbed for validating the logical consistency of our synthesis engine. Figure~\ref{fig:trajectory_vis} presents the visualization of these key timestamps.

\begin{figure*} 
    \centering
    \begin{subfigure}{\textwidth}
        \centering
        \includegraphics[width=0.95\textwidth]{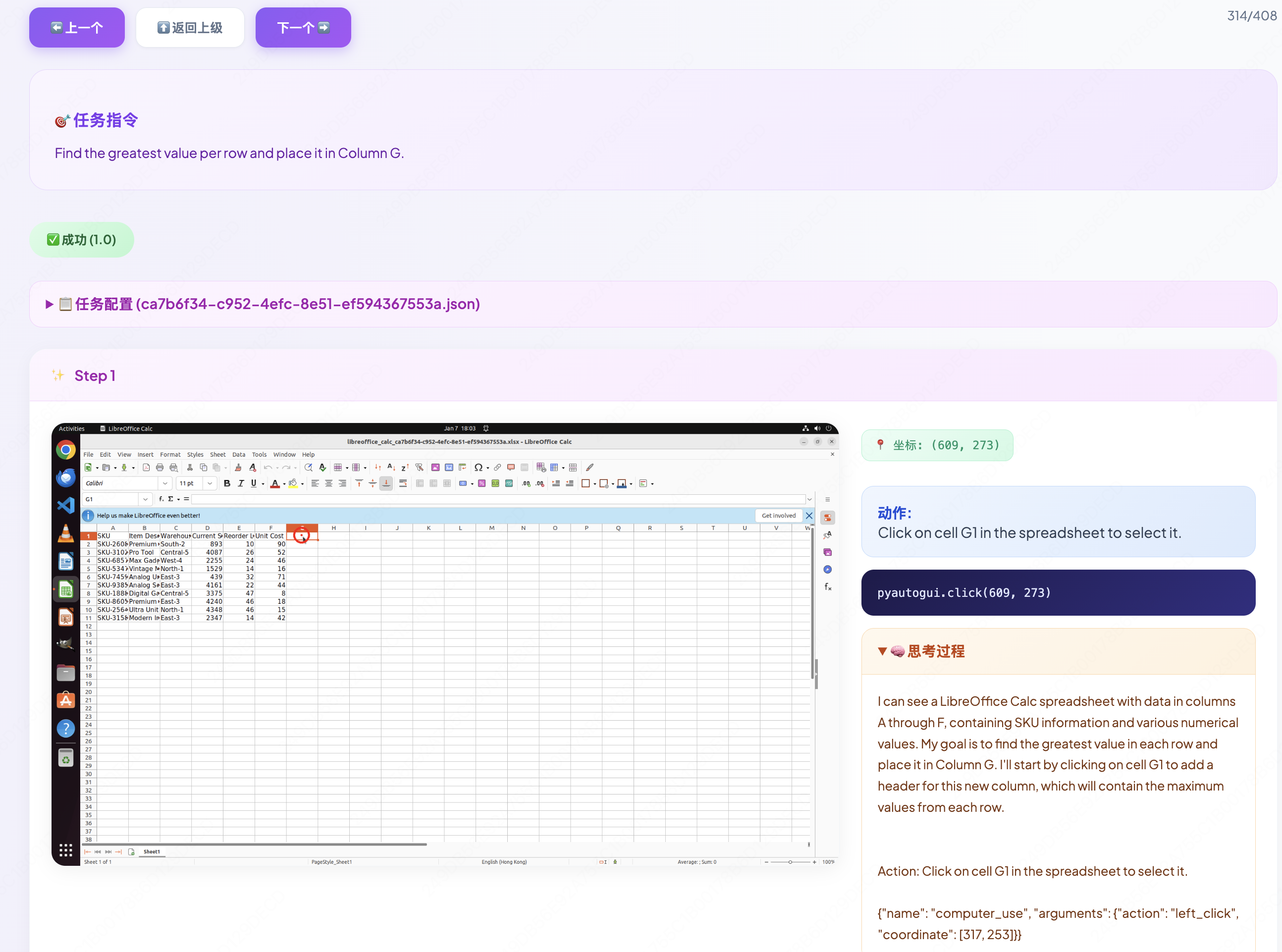}
        \caption{\textbf{Step 1: Goal Clarification ($t=1$).} The inspector visualizes the initial state. The reasoning panel displays the agent's explicit paraphrasing of the instruction ("Find the greatest value... place it in Column G"), validating the goal grounding mechanism in our synthetic data.}
        \label{subfig:step1}
    \end{subfigure}
    
    \vspace{10pt}
    \centerline{\Large $\downarrow$} 
    \vspace{10pt}

    \begin{subfigure}{\textwidth}
        \centering
        \includegraphics[width=0.95\textwidth]{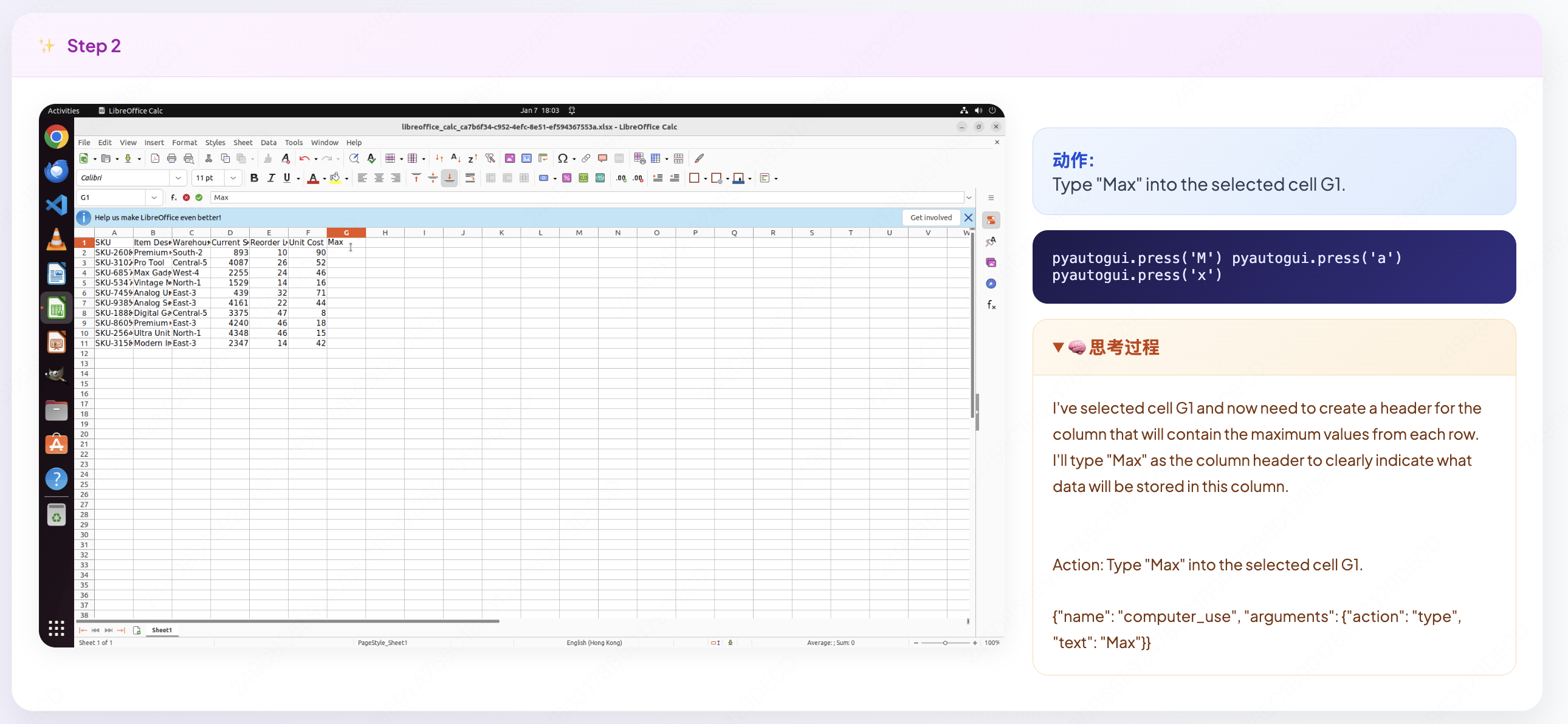}
        \caption{\textbf{Step 2: Text Entry ($t=1$).} The system captures the transition from planning to execution. The agent's reasoning ("Type 'Max'...") is perfectly aligned with the generated atomic action sequence (\texttt{press('M'), press('a'), press('x')}).}
        \label{subfig:step2}
    \end{subfigure}

    \caption{\textbf{Visualization of Synthesized Trajectory (Part I).} The EvoCUA Trajectory Inspector validates the logical consistency of synthetic training data: (a) Clear alignment between user instruction and agent planning; (b) Precise correspondence between reasoning and atomic keyboard actions.}
    \label{fig:trajectory_vis}
\end{figure*}

\begin{figure*}
    \ContinuedFloat 
    \centering
    
    \centerline{\Large $\vdots$} 
    \vspace{10pt}

    \begin{subfigure}{\textwidth}
        \centering
        \includegraphics[width=0.95\textwidth]{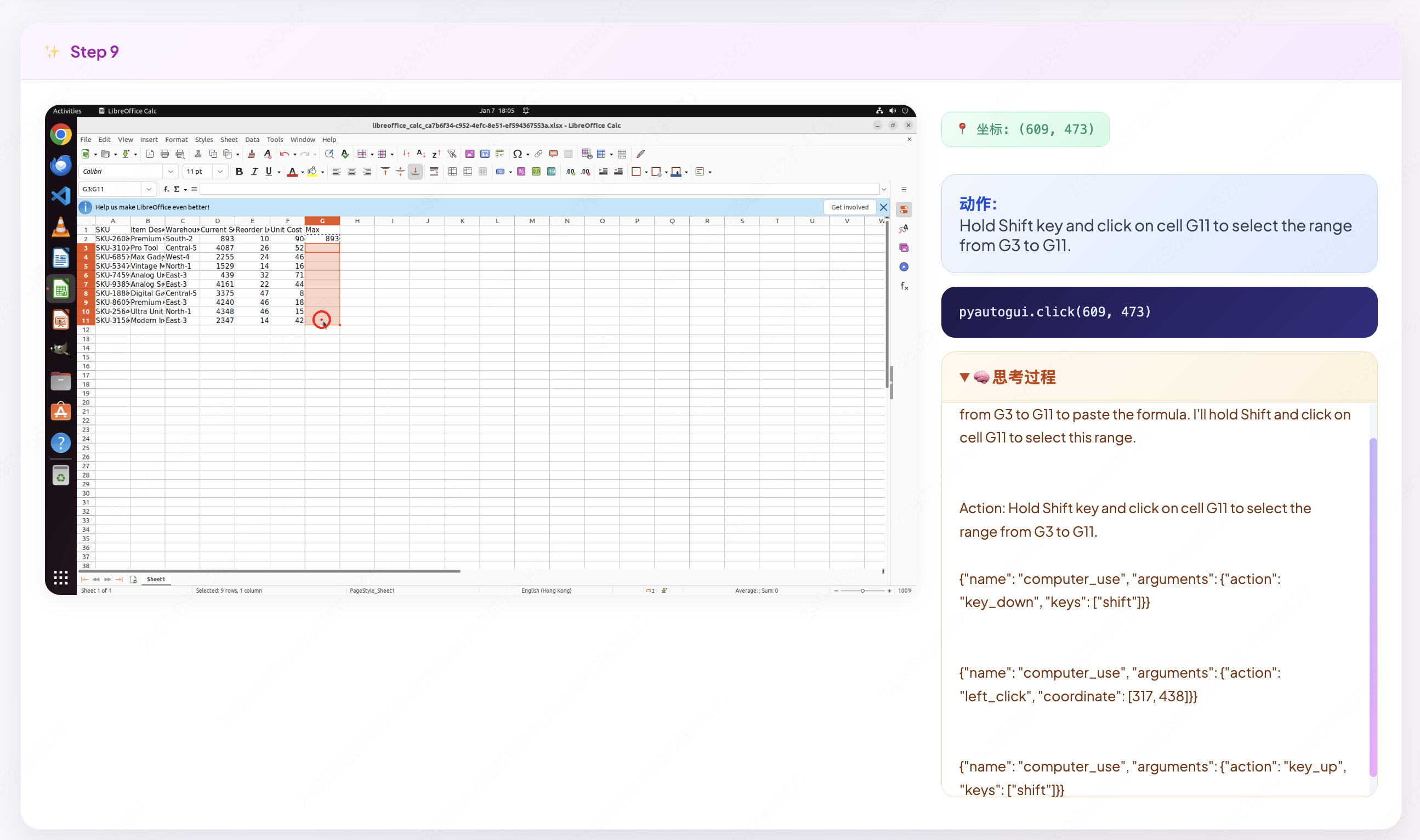}
        \caption{\textbf{Step 9: Stateful Interaction ($t=9$).} This view validates the \textit{Unified Action Space}. The synthetic ground truth requires a stateful operation (Shift-Select). The inspector confirms the agent correctly executes the \texttt{key\_down: shift} $\rightarrow$ \texttt{click} $\rightarrow$ \texttt{key\_up: shift} sequence.}
        \label{subfig:step9}
    \end{subfigure}

    \vspace{10pt}
    \centerline{\Large $\vdots$} 
    \vspace{10pt}

    \begin{subfigure}{\textwidth}
        \centering
        \includegraphics[width=0.95\textwidth]{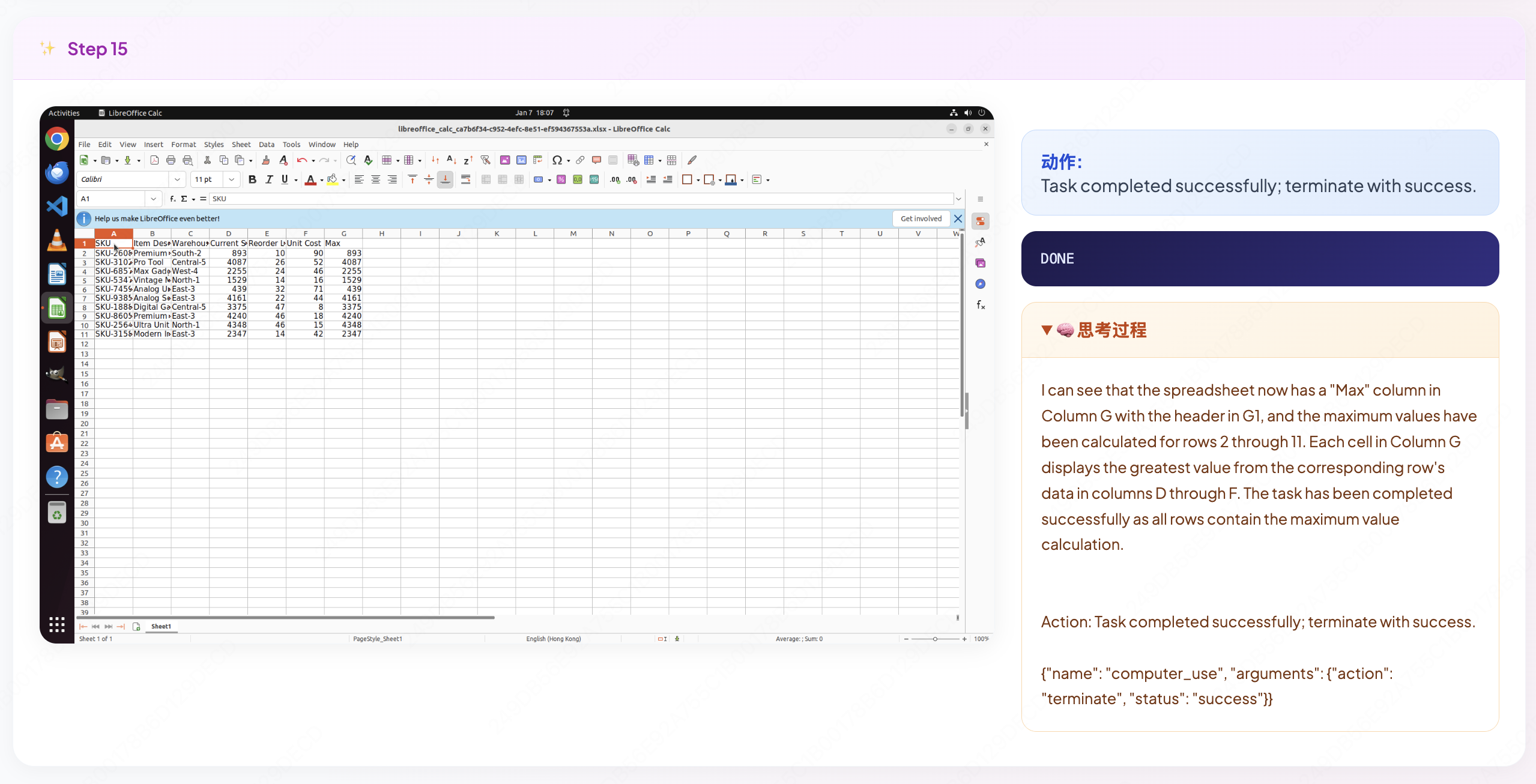}
        \caption{\textbf{Step 15: Verified Termination ($t=15$).} The final frame validates the \textit{Reasoning-Augmented Termination} schema. The tool highlights that the agent generates visual evidence ("I can see... Max column... calculated") to justify the successful termination status.}
        \label{subfig:step15}
    \end{subfigure}

    \caption{\textbf{Visualization of Synthesized Trajectory (Part II).} (c) Validation of complex stateful primitives (Shift+Click) essential for GUI manipulation. (d) Validation of the termination logic to ensure task completeness.}
\end{figure*}